\documentclass{article}

\usepackage{microtype}
\usepackage{graphicx}
\usepackage{subcaption}
\usepackage{booktabs} 

\usepackage{hyperref}
\usepackage{appendix}
\usepackage{titletoc}

\usepackage[preprint]{icml2026}

\usepackage{amsmath}
\usepackage{amssymb}
\usepackage{mathtools}
\usepackage{amsthm}
\usepackage[utf8]{inputenc}
\usepackage{booktabs}
\usepackage{multirow}
\usepackage{graphicx}
\usepackage[table]{xcolor}
\usepackage{lipsum} 
\usepackage{pifont}
\usepackage{cuted}
\definecolor{stdcolor}{HTML}{f2f3f5} 

\definecolor{twicolor}{HTML}{fff5e6}

\definecolor{gencolor}{HTML}{e0f7fa}   
\definecolor{ourscolor}{HTML}{f2f2f2}

\usepackage[capitalize,noabbrev]{cleveref}

\theoremstyle{plain}

\theoremstyle{definition}

\theoremstyle{remark}

\usepackage[textsize=tiny]{todonotes}

\NewDocumentCommand{\yafu}
{ mO{} }{\textcolor{cyan}{\textsuperscript{\textit{yafu}}\textsf{\textbf{\small[#1]}}}}

\begin{document}
  \twocolumn[\icmltitle{DiffThinker: Towards Generative Multimodal Reasoning with Diffusion Models}
  
\begin{icmlauthorlist}
\icmlauthor{Zefeng He}{shlab,nju}
\icmlauthor{Xiaoye Qu}{shlab}
\icmlauthor{Yafu Li}{shlab,cuhk}
\icmlauthor{Tong Zhu}{shlab}
\icmlauthor{Siyuan Huang}{shlab,sj}
\icmlauthor{Yu Cheng}{cuhk}
\end{icmlauthorlist}
\begin{center}
    \vspace{8pt}
    \textbf{Project Page:} \href{https://diffthinker-project.github.io}{\textcolor{blue}{https://diffthinker-project.github.io}}
    \vspace{-32pt}
\end{center}
]
\icmlaffiliation{shlab}{Shanghai AI Laboratory}
\icmlaffiliation{nju}{Nanjing University}
\icmlaffiliation{sj}{
Shanghai Jiao Tong University}
\icmlaffiliation{cuhk}{The Chinese University of Hong Kong}
\icmlcorrespondingauthor{Xiaoye Qu}{quxiaoye@pjlab.org.cn} 
\icmlcorrespondingauthor{Yu Cheng}{chengyu@cse.cuhk.edu.hk}

\printAffiliationsAndNotice{} 

\begin{strip}
        \centering
        \vspace{-10pt} 
        \begin{minipage}{0.33\textwidth}
            \centering
            \includegraphics[width=\linewidth]{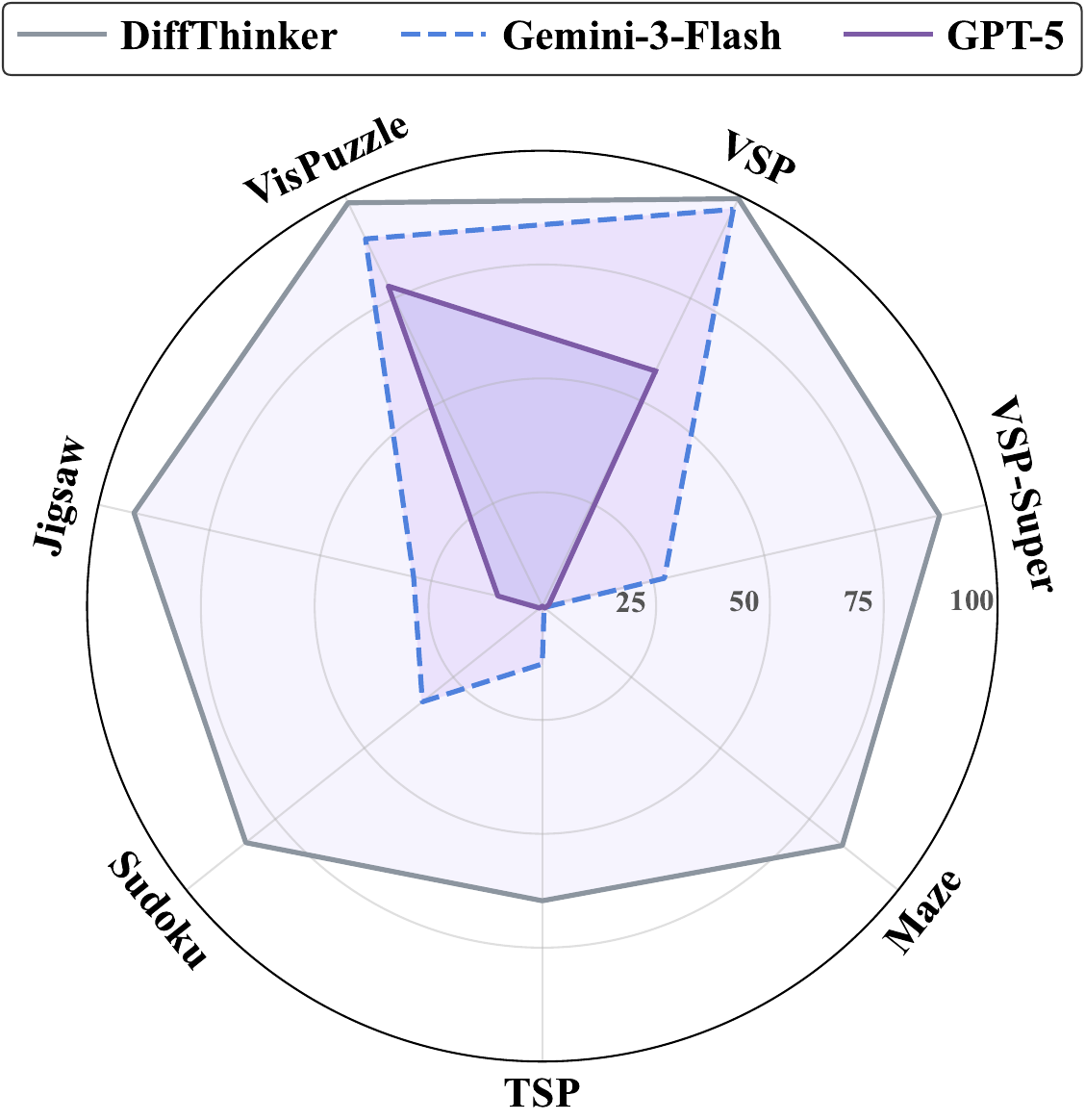}
            \vspace{-5pt}
            \centerline{\footnotesize (a) Overall performance.}
        \end{minipage}%
        \begin{minipage}{0.65\textwidth}
            \centering
            \includegraphics[width=\linewidth]{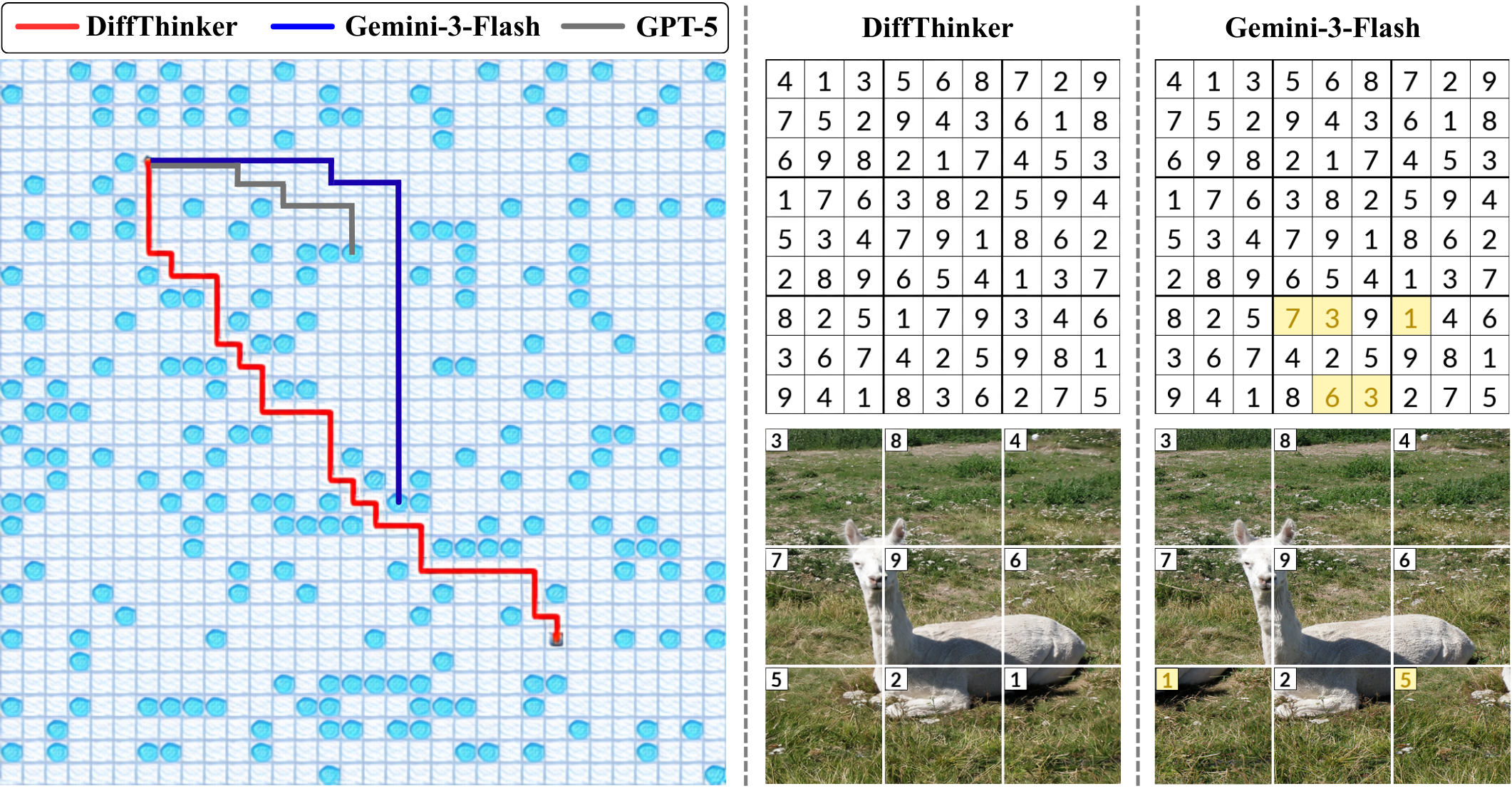}
            \vspace{-5.5pt}
            \centerline{\raisebox{1.0pt}{\footnotesize (b) Visualizations on VSP-Super, Sudoku and Jigsaw.}}
        \end{minipage}%
        
        \vspace{15pt}
    \captionof{figure}{(a) Quantitative results across seven tasks. (b) DiffThinker produces solution images directly, whereas baseline results are post-processed visualizations of textual outputs with errors highlighted. By reformulating reasoning as a native image-to-image generative task, DiffThinker achieves superior logical consistency and spatial precision in complex long-horizon, vision-centric reasoning tasks.}
        \label{fig:teaser}
        \vspace{20pt} 
\end{strip}

\begin{abstract}
While recent Multimodal Large Language Models (MLLMs) have attained significant strides in multimodal reasoning, their reasoning processes remain predominantly text-centric, leading to suboptimal performance in complex long-horizon, vision-centric tasks. 
In this paper, we establish a novel Generative Multimodal Reasoning paradigm and introduce DiffThinker, a diffusion-based reasoning framework.
Conceptually, DiffThinker reformulates multimodal reasoning as a native generative image-to-image task, achieving superior logical consistency and spatial precision in vision-centric tasks.
We perform a systematic comparison between DiffThinker and MLLMs, providing the first in-depth investigation into the intrinsic characteristics of this paradigm, revealing four core properties: efficiency, controllability, native parallelism, and collaboration. Extensive experiments across four domains (sequential planning, combinatorial optimization, constraint satisfaction, and spatial configuration) demonstrate that DiffThinker significantly outperforms leading closed source models including GPT-5 (+314.2\%) and Gemini-3-Flash (+111.6\%), as well as the fine-tuned Qwen3-VL-32B baseline (+39.0\%), highlighting generative multimodal reasoning as a promising approach for vision-centric reasoning.
\end{abstract}
    
\section{Introduction}

In recent years, Multimodal Large Language Models (MLLMs)~\cite{gemini3,gpt5,qwen3vl,gemini2.5} have achieved remarkable progress in multimodal understanding. The introduction of Chain-of-Thought (CoT) empowers these models with complex reasoning capabilities. Furthermore, Reinforcement Learning with Verifiable Reward \cite{deepseekmath,deepseekr1,zhang2025survey} has substantially enhanced the reasoning capabilities of MLLMs~\cite{wang2025vl}. Building upon these foundations, the emerging paradigm of ``Thinking with Image'' \cite{o3,zheng2025deepeyes,wang2025pixel,su2025thinking} enables MLLMs to interact with multimodal inputs iteratively, further pushing the boundaries of multimodal reasoning. 

Despite these advances, current MLLMs primarily rely on lengthy CoT for reasoning, resulting in uncontrollable generation and prohibitive latency~\cite{sui2025stopoverthinkingsurveyefficient,qu2025survey}. This inefficiency is further intensified by the multi-turn interactions inherent in the ``Thinking with Image'' paradigm. More importantly, these reasoning processes stay predominantly text-centric and struggle to track the changing state of visual information over long sequences, posing significant challenges for complex long-horizon, vision-centric tasks 
~\cite{vsp,maze}.

To address these limitations, in this paper, we introduce \textbf{DiffThinker}, and establish \textbf{Generative Multimodal Reasoning} as a novel paradigm that shifts the reasoning from symbolic space to visual space. 
Unlike MLLMs that typically conceptualize reasoning as a multimodal-to-text mapping, we propose to model multimodal reasoning directly as a generative image-to-image task with diffusion models. 
We conduct a systematic comparison between DiffThinker and MLLMs across a diverse set of challenging tasks, and provide the first in-depth investigation into the intrinsic characteristics of generative multimodal reasoning, revealing four core properties of DiffThinker:
\ding{172} \textbf{Efficient Reasoning:} It demonstrates superior efficiency in both training and inference, as well as higher accuracy, significantly outperforming RL-based MLLMs.
\ding{173} \textbf{Controllable Reasoning:} It provides stable and controllable inference costs, contrasting with the variable length CoT in MLLMs. 
\ding{174} \textbf{Native Parallel Reasoning:} It inherently explores multiple candidate solutions in parallel, progressively pruning invalid paths. 
\ding{175} \textbf{Collaborative Reasoning:} It acts as a partner with MLLMs, achieve performance surpassing either model alone.

To comprehensively evaluate the performance of DiffThinker, we conduct experiments across seven tasks in four domains including sequential planning, combinatorial optimization, constraint satisfaction, and spatial configuration. The results demonstrate that DiffThinker significantly outperforms state-of-the-art MLLMs, including GPT-5 (\textbf{+314.2\%}), Gemini-3-Flash (\textbf{+111.6\%}), and the Qwen3-VL-32B baseline fine-tuned on identical datasets (\textbf{+39.0\%}). 
Furthermore, we extend DiffThinker to the image-to-video generation paradigm for multimodal reasoning, and propose the DiffThinker-Video variant, demonstrating that video generation also exhibits inherent multimodal reasoning capabilities, and further highlight the effectiveness and efficiency of DiffThinker through comparative evaluations.

In summary, our contributions are threefold:
\begin{itemize}
     \item We propose DiffThinker and establish Generative Multimodal Reasoning as a novel paradigm, reformulating multimodal reasoning from text-centric symbolic mapping to a native image-to-image generative process.
     
    \item We perform a systematic comparison between DiffThinker and MLLMs across multiple domains and conduct the first investigation into the intrinsic characteristics of this generative multimodal reasoning paradigm,
    revealing four core properties: efficiency, controllability, native parallelism, and collaboration.
    
    \item Extensive experiments on seven tasks demonstrate {DiffThinker} significantly outperforms SOTA MLLMs including GPT-5 (+314.2\%) and Gemini-3-Flash (+111.6\%), revealing generative multimodal reasoning as a promising approach for vision-centric reasoning.
\end{itemize}

\begin{figure*}[t]
    \centering
    \includegraphics[width=1\linewidth]{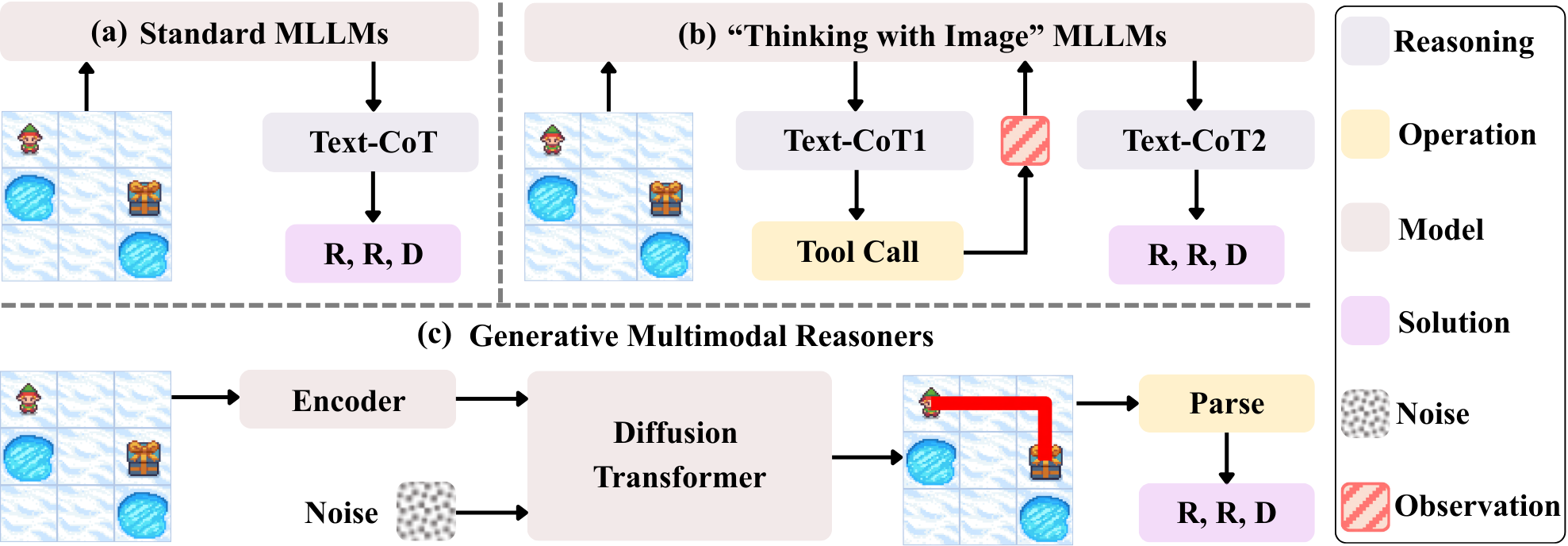}
    \caption{{Overview of different multimodal reasoning paradigms.} (a) Standard MLLMs map inputs directly to symbolic solutions. (e.g., `R' and `D' representing `Right' and `Down' actions) (b) ``Thinking with Images'' MLLMs interact with multimodal inputs through iterative tool calls. (c) DiffThinker reformulates multimodal reasoning as a direct generative image-to-image task, where solutions are produced in visual space and then parsed to symbolic solutions to ensure a fair comparison.}
    \label{fig:reformulation}
\end{figure*}

\section{Related Works}
\subsection{Multimodal Reasoning}
Reinforcement Learning with Verifiable Reward~\cite{deepseekr1,deepseekmath} has significantly enhanced LLM reasoning, and is rapidly extending to MLLMs~\cite{huang2025vision,shen2025vlm,liu2025visual,huang2025spotlight,he2025videossr,wang2025vl,shen2025satori}. However, existing paradigms remain predominantly text-centric, which hinders performance in vision-centric tasks.
Advancing this frontier, the paradigm of ``Thinking with Image''~\cite{o3} introduces a mechanism for models to engage in multi-turn visual interactions during the reasoning process. While earlier approaches~\cite{su2025openthinkimg,zheng2025deepeyes,wang2025pixel,hong2025deepeyesv2,zhang2025thyme,lai2025mini} relied on tool calls or code execution for image manipulation, recent works~\cite{yang2025machine,xu2025visual,du2025revisiting,zhang2025latent,wang2025monet,chen2025reasoning,qin2025chain,thinkmorph} have shifted toward generating native images or latent visual tokens. Nevertheless, the underlying architectures of these methods remain rooted in autoregressive MLLMs, leading to limited performance in complex long-horizon, vision-centric tasks. 
 
Building upon the success of ``Thinking with Image,'' the ``Thinking with Video'' paradigm enhances reasoning by enabling models to interact with video content through multi-turn tool invocation~\cite{zhang2025thinking,he2025framethinker,yan2025videochat,xie2025video}. This concept has recently advanced to performing multimodal reasoning directly through video generation~\cite{wiedemer2025video,tong2025thinking,yang2025reasoning,luo2025v,liu2025can,guo2025video,miniveo3reasoner}. However, these studies predominantly focus on benchmarking closed source models~\cite{veo3,sora} with undisclosed internal reasoning mechanisms. Furthermore, video generation itself entails prohibitive computational costs. Diverging from this, DiffThinker establishes image generation as a more efficient paradigm.

\subsection{Diffusion Models}
Diffusion models have emerged as the dominant framework for generative modeling. Early research~\cite{sohl2015deep,song2019generative,ho2020denoising,song2020denoising,ho2022classifier} laid the theoretical foundations of this paradigm. Subsequently, flow-based methodologies~\cite{lipman2022flow,liu2022flow,albergo2022building} have further advanced the field. The integration of latent diffusion models~\cite{rombach2022high}, diffusion transformers~\cite{peebles2023scalable}, and multimodal diffusion transformers~\cite{mmdit} has established the current mainstream for generative modeling, paving the way for diverse downstream applications.

While one prominent direction of research concentrates on high-fidelity image~\cite{rombach2022high,ramesh2022hierarchical,saharia2022photorealistic} and video~\cite{ho2022video,wan,brooks2024video} generation, other applications extend to specialized tasks~\cite{avdeyev2023dirichlet,ubukata2024diffusion,pogodzinski2025spatial,graikos2022diffusion,li2024fast,li2023t2t}, such as Sudoku~\cite{wewer2025spatial}, geometry~\cite{goren2025visual}, and the Traveling Salesperson Problem~\cite{sun2023difusco}. Diverging from these specialized approaches, we focus on multimodal reasoning and introduce DiffThinker, establishing Generative Multimodal Reasoning as a novel paradigm. Also, unlike prior methods that typically require task-specific architectures and training from scratch, DiffThinker enables rapid adaptation to diverse multimodal reasoning tasks by formulating them as a unified generative process in visual space.

\begin{figure*}[t]
\centering
\includegraphics[width=\textwidth]{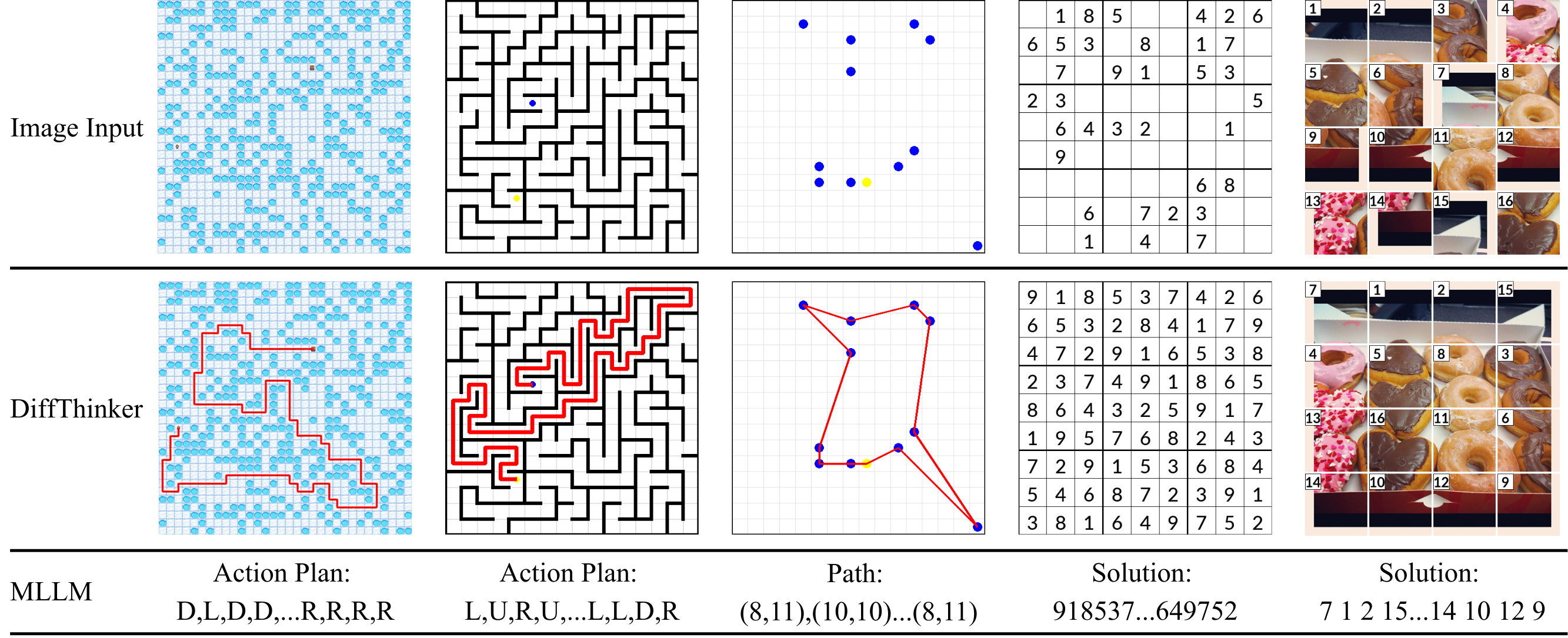}
\caption{\textbf{Main tasks.} Each column represents a specific task. The first row displays the image input. The second row shows the results generated by DiffThinker. The third row presents the outputs from the MLLM baseline.}
\label{fig:task_overview}
\end{figure*}

\section{Generative Multimodal Reasoning}
\subsection{Problem Reformulation}

In this work, we introduce DiffThinker, a generative multimodal reasoner that innovatively reformulates multimodal reasoning as an image-to-image task, as illustrated in Figure~\ref{fig:reformulation}. To clarify the paradigm shift, we define and formalize three distinct reasoning paradigms.

\textbf{Standard MLLMs: Multimodal-to-Text.} Standard MLLMs model the reasoning process as a sequential mapping in the symbolic space. Given a visual input $x \in \mathcal{X}$ and a textual instruction $c \in \mathcal{T}$, the process is defined as:
\begin{equation}
    f_{\text{Std}}(x, c) \to z \to y,
\end{equation}
where $z$ represents the textual reasoning trace (e.g., Chain-of-Thought) and $y \in \mathcal{Y}$ is the final solution. Despite their success, the reasoning process remains text-centric, often leading to suboptimal performance in vision-centric tasks.

\textbf{``Thinking with Image'' MLLMs: Iterative Interaction.} This paradigm enhances reasoning by enabling models to interact with multimodal inputs to generate intermediate results via tool calls. The process is formulated as an interleaved sequence of reasoning, tool call, and observation:
\begin{equation}
    f_{\text{TwI}}(x, c) \to \{(z_1, t_1, o_1), \dots, (z_n, t_n, o_n)\} \to y,
\end{equation}
where $z_i$ denotes the $i$-th reasoning step, $t_i$ represents the tool call, $o_i$ is the corresponding intermediate visual observation, and $y \in \mathcal{Y}$ signifies the final solution. Although this paradigm incorporates essential visual feedback, its inherent reliance on iterative multi-turn loops and the associated computational overhead pose significant challenges for scaling to complex long-horizon, vision-centric tasks.

\textbf{DiffThinker: Multimodal-to-Image.} 
Unlike MLLMs, which primarily reason within symbolic space, DiffThinker shifts the reasoning process into visual space through a direct multimodal-to-image transformation. In this generative multimodal reasoning paradigm, the model functions as a generator $G$ that directly produces a solution image $x_{sol}$ from the visual input $x$ and the textual instruction $c$:
\begin{equation}
    G(x, c) \to x_{sol} \in \mathcal{X},
\end{equation}
where $x_{sol}$ visually encapsulates the reasoning trajectory and solution. To facilitate comparison with the symbolic ground-truth, we introduce a parsing function $\Psi: \mathcal{X} \to \mathcal{Y}$ to map the solution image to symbolic space:
\begin{equation}
    y_{parsed} = \Psi(x_{sol}).
\end{equation}
Rather than relying on MLLMs to judge whether a solution image conforms to the textual ground-truth, our parsing mechanism ensures a fair comparison across different paradigms and precludes potential answer leakage.

\begin{table*}[t]
\centering
\caption{\textbf{Comprehensive Results across All Tasks.} We evaluate models across a total of four domains including Sequential Planning (VSP, VSP-Super, and Maze), Combinatorial Optimization (TSP), Constraint Satisfaction (Sudoku), and Spatial Configuration (Jigsaw and VisPuzzle). Evaluation is conducted on varying difficulty levels, defined by grid size for Sequential Planning and Jigsaw, number of cities for TSP, and number of given clues for Sudoku. ``N/A'' denotes vanilla models without training. The Avg column represents the grand mean calculated from individual task averages.}
\label{tab:merged_results}
\resizebox{\textwidth}{!}{
\begin{tabular}{l|c|cccccc|cc|ccc|ccc|ccc|ccc|c|c}
\toprule
\multirow{2}{*}{\textbf{Model}} & \multirow{2}{*}{\textbf{Setting}} & \multicolumn{6}{c|}{\textbf{VSP}} & \multicolumn{2}{c|}{\textbf{VSP-Super}} & \multicolumn{3}{c|}{\textbf{Maze}} & \multicolumn{3}{c|}{\textbf{TSP}} & \multicolumn{3}{c|}{\textbf{Sudoku}} & \multicolumn{3}{c|}{\textbf{Jigsaw}} & \multirow{2}{*}{\textbf{VisP.}}  & \multirow{2}{*}{\textbf{Avg}}\\ \cline{3-22}
 &  & \textbf{3} & \textbf{4} & \textbf{5} & \textbf{6} & \textbf{7}  & \textbf{8}  & \textbf{16} & \textbf{32} & \textbf{8} & \textbf{16} & \textbf{32} & \textbf{12} & \textbf{15} & \textbf{18}  & \textbf{45} & \textbf{40} & \textbf{35} & \textbf{2} & \textbf{3} & \textbf{4} & \\ \midrule
\rowcolor{stdcolor} \multicolumn{24}{c}{\textbf{Closed Source MLLMs}} \\ \midrule
Gemini-3-Flash & N/A & 100 & 100 & 100 & 99 & 83 & 98 & 52 & 3 & 0 &0 & 0 & 25 & 9 & 4 & 69 & 29 & 3 & 71 & 16 & 0 & 89.5& 41.3 \\ 
GPT-5 & N/A & 99 & 70 & 67 & 43 & 36 & 29 & 3 &0 & 2 & 0 & 0 & 0 & 0 & 0 & 2 & 0 & 0 & 30 & 0 & 0 & 78.0 &21.1\\
\midrule
\rowcolor{twicolor} \multicolumn{24}{c}{\textbf{Open Source MLLMs}} \\ \midrule
\multirow{3}{*}{Qwen3-VL-8B} & N/A & 64 & 46 & 33 & 21 & 12 & 21 & 1 & 0 & 0 & 0 & 0 & 0 & 0 & 0 & 0 & 0 & 0 & 7 & 0 & 0 & 28.0&9.1 \\
 & SFT & 99 & 96 & 98 & 96 & 92 & 86 & 61 & 8 & 53 & 37 & 0 & 59 & 60 & 43 & 30 & 17 & 2 & 95 & 56 & 9 & 78.8 &51.6\\
 & GRPO & 91 & 70 & 70 & 31 & 34 & 24 & 0 & 0 & 0 &0 & 0 & 0 & 0 & 0 & 0 & 0 & 0 & 6 & 0 & 0 & 28.0&11.9 \\ \midrule
\multirow{3}{*}{Qwen3-VL-32B} & N/A & 75 & 51 & 47 & 25 & 23 & 26 & 0 & 0 & 0 & 0 & 0 & 0 & 0 & 0 & 0 & 0 & 0 & 9 & 0 & 0 & 29.5&10.5 \\
 & SFT & 96 & 99 & 98 & 100 & 99 & 90 & 85 & 21 & 91 & 57 & 3 & 69 & 59 & 52 & 32 & 22 & 2 & 97 & 72 & 28 & 95.8&62.9 \\
 & GRPO & 99 & 90 & 95 & 69 & 73 & 58 & 1 & 0 & 0 & 0 & 0 & 0 & 0 & 0 & 3 & 1 & 0 & 64 & 4 & 0 & 83.0&26.9 \\ \midrule
\rowcolor{gencolor} \multicolumn{24}{c}{\textbf{Generative Multimodal Reasoners}} \\ \midrule
Qwen-Image-Edit-2509 & N/A & 33 & 36 & 22 & 12 & 11 & 7 & 0 & 0 & 0 & 0 & 0 & 0 & 0 & 0 & 0 & 0 & 0 & 0 & 0 & 0 & 7.5&4.0 \\
\rowcolor{ourscolor} \textbf{DiffThinker (Ours)} & Flow Matching & \textbf{99} & \textbf{100} & \textbf{98} & \textbf{99} & \textbf{100} & \textbf{100} & \textbf{96} & \textbf{83} & \textbf{100} & \textbf{97} & \textbf{56} & \textbf{74} & \textbf{62} & \textbf{58} & \textbf{98} & \textbf{95} & \textbf{57} & \textbf{99} & \textbf{97} & \textbf{80} & \textbf{98.3} &\textbf{87.4}\\ \midrule
Qwen-Image-Edit-2511 & N/A & 50 & 55 & 44 & 16 & 23 & 23 & 0 & 0 & 0 & 0 & 0 & 0 & 0 & 0 & 0 & 0 & 0 & 0 & 0 & 0 & 11.5 &6.7\\ 
\rowcolor{ourscolor} \textbf{DiffThinker++ (Ours)} & Flow Matching & \textbf{100} & \textbf{100} & \textbf{100} & \textbf{98} & \textbf{100} & \textbf{100} & \textbf{99} & \textbf{80} & \textbf{100} & \textbf{100} & \textbf{65} & \textbf{76} & \textbf{72} & \textbf{59} & \textbf{97} & \textbf{94} & \textbf{55} & \textbf{99} & \textbf{98} & \textbf{80} & \textbf{98.8} &\textbf{88.5}\\ \bottomrule
\end{tabular}
}
\end{table*}

\subsection{Flow Matching}

DiffThinker is implemented based on Qwen-Image-Edit~\cite{wu2025qwen}. 
Mathematically, it employs Flow Matching~\cite{lipman2022flow, liu2022flow, albergo2022building} as the theoretical framework to approximate the velocity field that transforms noise into the data distribution, ensuring stable learning dynamics via Ordinary Differential Equations (ODEs). 
Architecturally, the model leverages a Multimodal Diffusion Transformer (MMDiT)~\cite{mmdit} to capture intricate cross-modal dependencies.
For efficiency, these generative processes are performed within the latent space of a Variational Autoencoder (VAE)~\cite{vae}.

\paragraph{Training.} 
Formally, let $y$ denote the ground-truth image. The data latent $x_0$ is obtained by encoding $y$ through the VAE encoder $\mathcal{E}$, i.e., $x_0 = \mathcal{E}(y)$. A random noise vector $x_1$ is sampled from the standard multivariate normal distribution, $x_1 \sim \mathcal{N}( \mathbf{0},  \mathbf{I})$. To incorporate multimodal task constraints, the conditioning latent $h$ is derived from the MLLM $\phi$ given the user instruction $S$ (comprising text and visual inputs), such that $h = \phi(S)$.
During training, a timestep $t$ is sampled from a logit-normal distribution with $t \in [0, 1]$. The intermediate latent variable $x_t$ is constructed via linear interpolation between the data $x_0$ and noise $x_1$:
\begin{equation}
\label{eq:fm}
    x_t = t x_0 + (1 - t) x_1.
\end{equation}
Consequently, the target velocity field $v_t$ driving the flow from noise to data is defined as:
\begin{equation}
    v_t = \frac{d x_t}{dt} = x_0 - x_1 .
\end{equation}
The MMDiT-based vector field $v_\theta$ is trained to predict this target velocity $v_t$. The training objective is formulated as the mean squared error (MSE):
\begin{equation}
    \mathcal{L}_{FM} = \mathbb{E}_{t, x_0, x_1} \left[ \| v_\theta(x_t, t, h) - (x_0 - x_1) \|^2 \right].
\end{equation}

\paragraph{Inference.} 
During inference, DiffThinker performs reasoning by solving the ODE defined by the learned velocity field $d x_t = v_\theta(x_t, t, h) dt$. From initial noise $x_{t=0} = x_1$, the model numerically integrates the flow to recover the solution latent $x_{t=1} \approx x_0$. Implementing a first-order Euler solver with a step size $\Delta t = 1/T$, the update rule is:
\begin{equation}
    x_{t+\Delta t} = x_t + \Delta t \cdot v_\theta(x_t, t, h).
\end{equation}
After $T$ steps, the final latent $x_{t=1}$ (which approximates the data distribution) is decoded back to pixel space via the VAE decoder to yield the visual solution: $y_{sol} = \mathcal{D}(x_{t=1})$.

\subsection{Task Formulation}

To systematically verify the efficacy of DiffThinker within the proposed generative reasoning paradigm, we select tasks based on three perspectives. First, we target complex long-horizon, vision-centric tasks that fundamentally rely on visual perception. 
Second, we prioritize tasks offering controllable and scalable difficulty levels, which facilitates a precise exploration of the model's capability boundaries. Third, we specifically select tasks featuring high structural parseability, such as grid-based configurations. Since the evaluation of DiffThinker involves parsing generated visual solutions into symbolic formats, this criterion ensures an objective assessment against ground-truth labels. Accordingly, our tasks contain five distinct classes as detailed below.

\textbf{Visual Spatial Planning (VSP)~\cite{vsp} and VSP-Super.} VSP evaluates perception and reasoning capabilities in spatial planning scenarios. We focus on its FrozenLake subset due to its parseability. 
Moreover, we introduce VSP-Super, which expands the environment scale. As illustrated in the first column of Figure~\ref{fig:task_overview}, the model must navigate a grid-based frozen lake while avoiding holes. DiffThinker generates a continuous visual trajectory rendered as a red line. Conversely, MLLMs produce text-based action plans. We formalize these challenges as sequential planning tasks.

\textbf{Maze~\cite{maze}.} 
This task involves longer routes than VSP, increasing navigation complexity. As illustrated in the second column of Figure~\ref{fig:task_overview}, the model must identify a path avoiding walls between cells. 
DiffThinker renders a trajectory from the yellow start to the blue target. Conversely, MLLMs output an action plan via a series of text tokens. We categorize this as a sequential planning task.

\textbf{Traveling Salesperson Problem (TSP)~\cite{tsp}.} This task requires solving the Traveling Salesperson Problem on a 2D plane, aiming to identify the shortest path visiting every city. As depicted in the third column of Figure~\ref{fig:task_overview}, the problem is visualized by a yellow start dot and blue city dots. DiffThinker generates a geometric path connecting all nodes into a closed loop. In contrast, MLLMs provide numerical coordinates to represent the order. This is classified as a combinatorial optimization problem.

\begin{figure*}[t]
    \centering
    \includegraphics[width=1.0\linewidth]{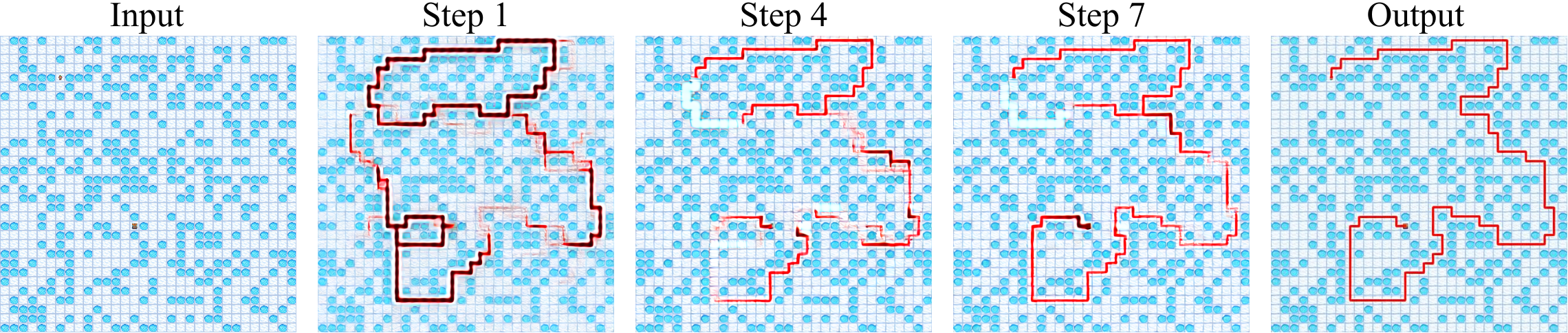}
    \caption{\textbf{DiffThinker as a native parallel reasoner.} Visualization of the native parallel reasoning process in DiffThinker. The model explores multiple candidate paths simultaneously in the early stages and iteratively refines them into a single valid trajectory.}
    \label{fig:parallel_reasoning}
\end{figure*}

\textbf{Sudoku.} In this task, the model must fill in missing digits while adhering to Sudoku constraints. As shown in the fourth column of Figure~\ref{fig:task_overview}, DiffThinker generates a completed grid with all empty cells populated. Conversely, MLLMs provide a text-based numerical sequence. This challenge is classified as a constraint satisfaction task.

\textbf{Jigsaw and VisPuzzle~\cite{thinkmorph}.} The Jigsaw task centers on spatial configuration and visual perception. As illustrated in the final column of Figure~\ref{fig:task_overview}, the input consists of shuffled patches, each numerically labeled to facilitate automated parsing. DiffThinker reconstructs these patches into a globally consistent image. In contrast, MLLMs produce a sequence of indices representing the restoration order. We also introduce VisPuzzle~\cite{thinkmorph}, which serves as a simplified benchmark for puzzle reconstruction. These challenges are categorized as spatial configuration tasks.

\section{Experiments}
\textbf{Experimental Setup.} 
DiffThinker is built upon Qwen-Image-Edit-2509~\cite{wu2025qwen}, utilizing a 20B MMDiT~\cite{mmdit}. Additionally, we implement DiffThinker++ based on the updated Qwen-Image-Edit-2511 for main results (Table~\ref{tab:merged_results}), whereas all subsequent analysis and ablation studies are conducted using DiffThinker. Following previous works~\cite{xu2025visual,miniveo3reasoner,yang2025reasoning}, we train independent models for VSP/VSP-Super, Maze, TSP, Sudoku, and Jigsaw, respectively. We also fine-tune Qwen3-VL baselines on identical datasets for a direct comparison. VisPuzzle serves as an out-of-distribution task for puzzle reconstruction. Evaluation is conducted on varying difficulty levels, as shown in Table~\ref{tab:merged_results}. Details are provided in Appendix~\ref{sec:training_details}.

\subsection{Main Results}

\begin{figure}[t]
    \centering
    \begin{subfigure}[b]{0.49\columnwidth}
        \centering
        \includegraphics[width=\textwidth]{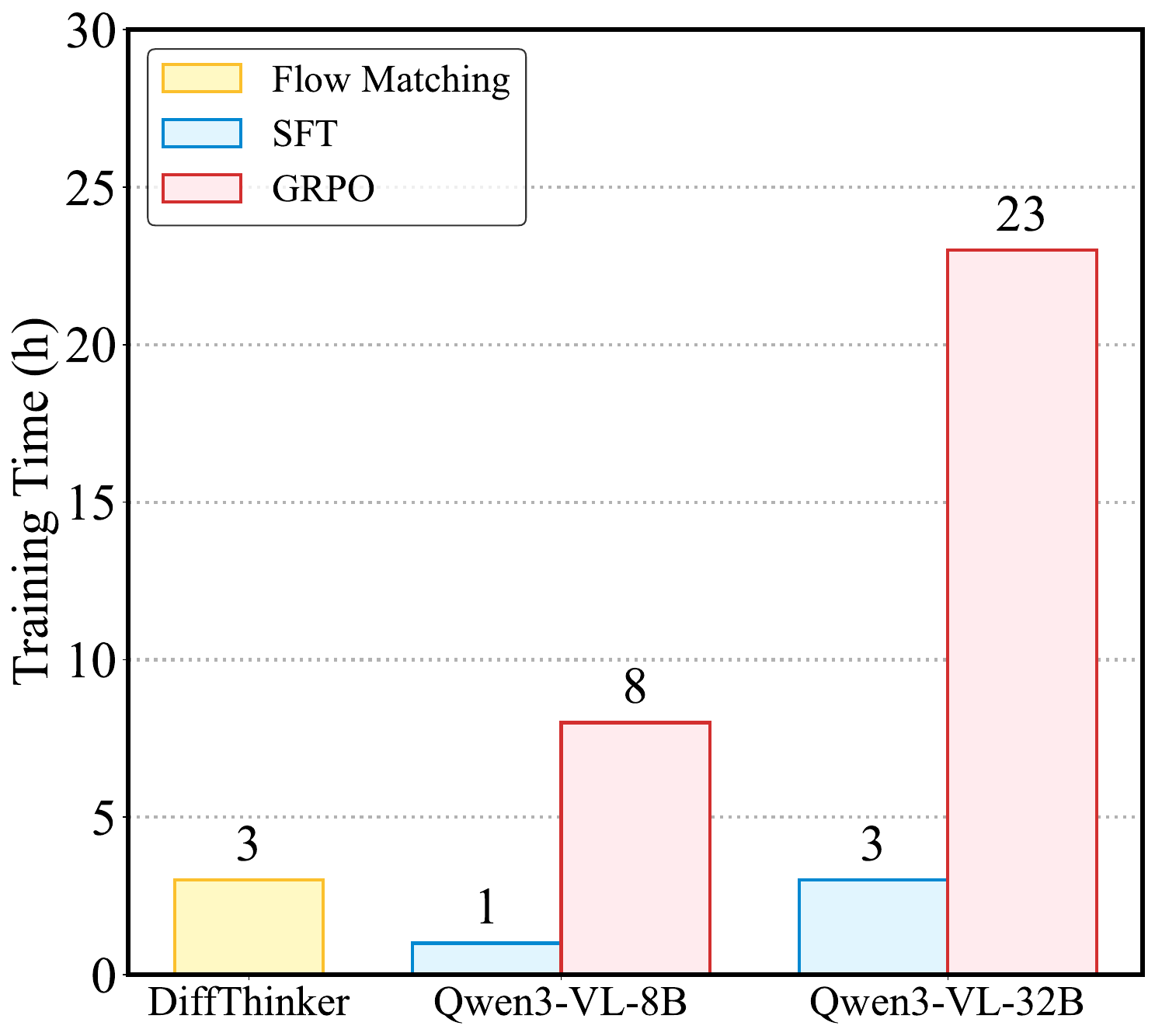}
        \caption{Training Time (h)}
        \label{fig:train_time}
    \end{subfigure}
    \hfill
    \begin{subfigure}[b]{0.49\columnwidth}
        \centering
        \includegraphics[width=\textwidth]{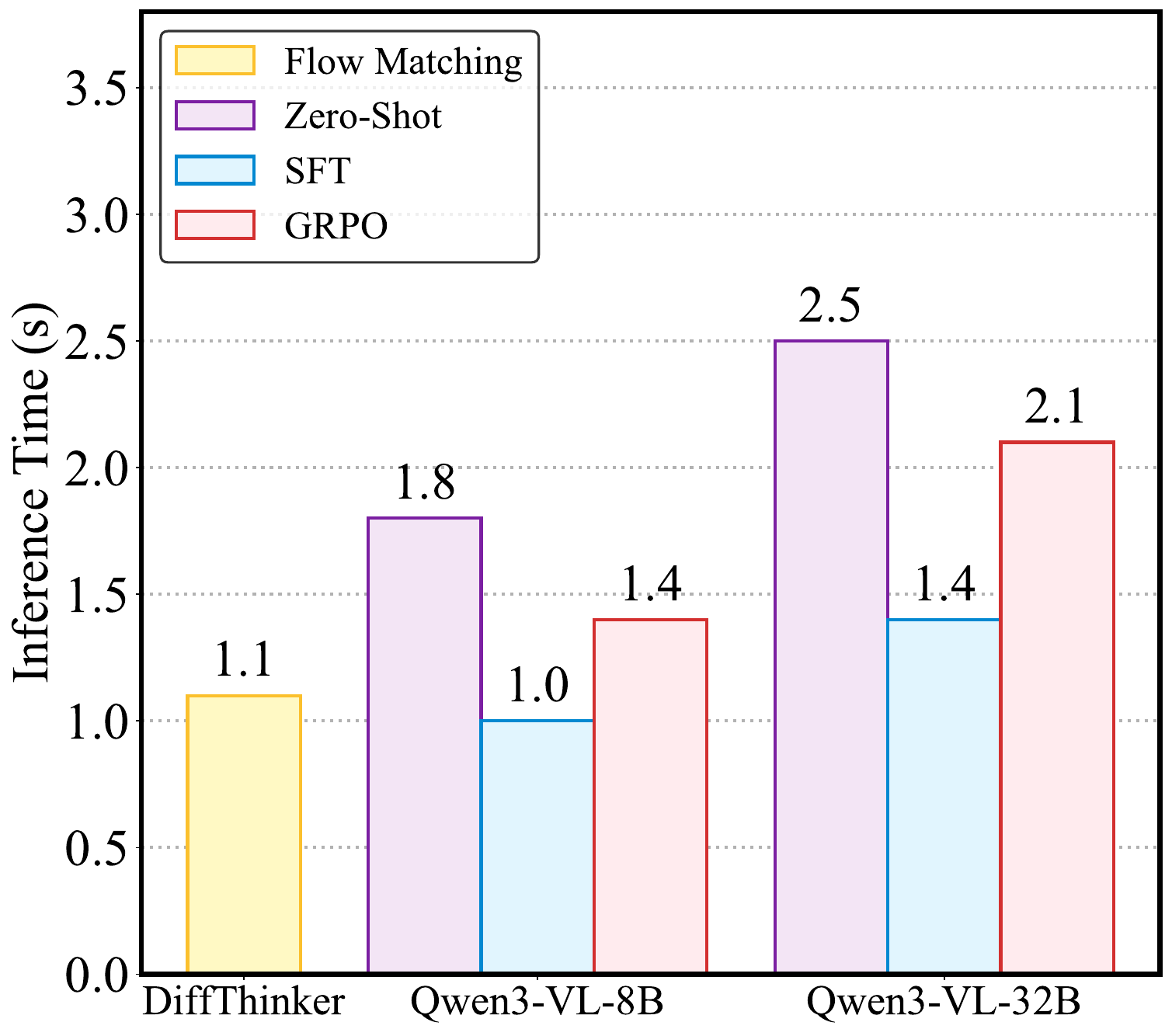}
        \caption{Inference Time (s)}
        \label{fig:infer_time}
    \end{subfigure}
    \caption{\textbf{Computational efficiency analysis}. (a) compares training duration in hours, and (b) shows inference latency in seconds.}
    \label{fig:efficiency_metrics}
\end{figure}

\textbf{DiffThinker as an Extraordinary Multimodal Reasoner.} 
As illustrated in Table~\ref{tab:merged_results}, DiffThinker achieves state-of-the-art performance across seven challenging tasks in four domains. Specifically, our approach drastically surpasses  GPT-5 (+314.2\%), Gemini-3-Flash (+111.6\%), and the fine-tuned Qwen3-VL-32B (+39.0\%) with fewer parameters.

Across all evaluated domains, DiffThinker demonstrates a clear advantage over traditional MLLMs. In sequential planning tasks such as VSP, VSP-Super, and Maze, the performance of MLLMs decays rapidly as task complexity scales, whereas DiffThinker maintains high accuracy through generative reasoning. In spatial configuration tasks including Jigsaw and VisPuzzle, the model achieves near-perfect performance, while similarly delivering exceptional results in combinatorial optimization (TSP) and constraint satisfaction (Sudoku). These results underscore that our generative multimodal reasoning paradigm provides a more robust foundation for multimodal reasoning than that of traditional MLLMs in long-horizon, vision-centric tasks.

\subsection{Discussion and Observation}

\textbf{DiffThinker as an Efficient Reasoner.} To quantitatively assess the computational overhead of DiffThinker relative to standard MLLMs, we conduct experiments to measure both training and inference durations on a cluster of eight NVIDIA H200 GPUs. We report training durations of VSP/VSP-Super, and the average inference latency of VSP-Super level-16 per reasoning instance. 

As illustrated in Figure~\ref{fig:efficiency_metrics}(a), DiffThinker maintains a highly competitive training efficiency. Its training duration is nearly identical to Qwen3-VL-32B (SFT) baseline and is substantially lower than the overhead of GRPO~\cite{deepseekmath}, a reinforcement learning paradigm currently widely adopted for multimodal reasoning. 
Regarding inference speed, as illustrated in Figure~\ref{fig:efficiency_metrics}(b), DiffThinker exhibits a highly competitive latency of 1.1s, which is comparable to Qwen3-VL-8B (SFT) baseline (1.0s) and faster than Qwen3-VL-32B (SFT) model (1.4s). 
This result underscores the inherent inference efficiency of our generative reasoning paradigm.

\textbf{DiffThinker as a Controllable Reasoner.} 
DiffThinker establishes a controllable reasoning paradigm by reformulating tasks into a fixed-step generative process. By employing an Euler solver with a predefined number of steps, the model ensures a deterministic computational budget which is invariant to the task's logical complexity. In contrast, MLLMs are plagued by unpredictable inference durations. Their autoregressive nature often leads to fluctuating latency caused by verbose Chain-of-Thought or repetitive output collapse, resulting in significantly longer average inference times, as shown in Figure~\ref{fig:efficiency_metrics}(b). Moreover, unlike MLLMs where imposing token limits risks premature truncation, the controllable generation of DiffThinker guarantees both execution stability and the derivation of reliable solutions.

 \textbf{DiffThinker as a Native Parallel Reasoner.} Unlike MLLMs, which execute reasoning sequentially and often require explicit reflection or backtracking to rectify early errors, DiffThinker possesses an inherent capacity for native parallel reasoning. To visualize the progressive reasoning process, we estimate the clean data latent at each intermediate timestep by projecting the current state back to the data manifold and decoding it into pixel space.
As illustrated in Figure~\ref{fig:parallel_reasoning}, during the initial reasoning stages (e.g., Step 1), DiffThinker avoids premature commitment to a single path, instead exploring multiple candidate trajectories across the grid in parallel. 
Through successive iterations, the model simultaneously evaluates global constraints and environmental obstacles to prune invalid routes, progressively consolidating its focus onto the most plausible path and eventually converging to an optimal solution.

\begin{figure}[t]
    \centering
    \begin{minipage}{0.38\columnwidth}
        \centering
        \includegraphics[width=\textwidth]{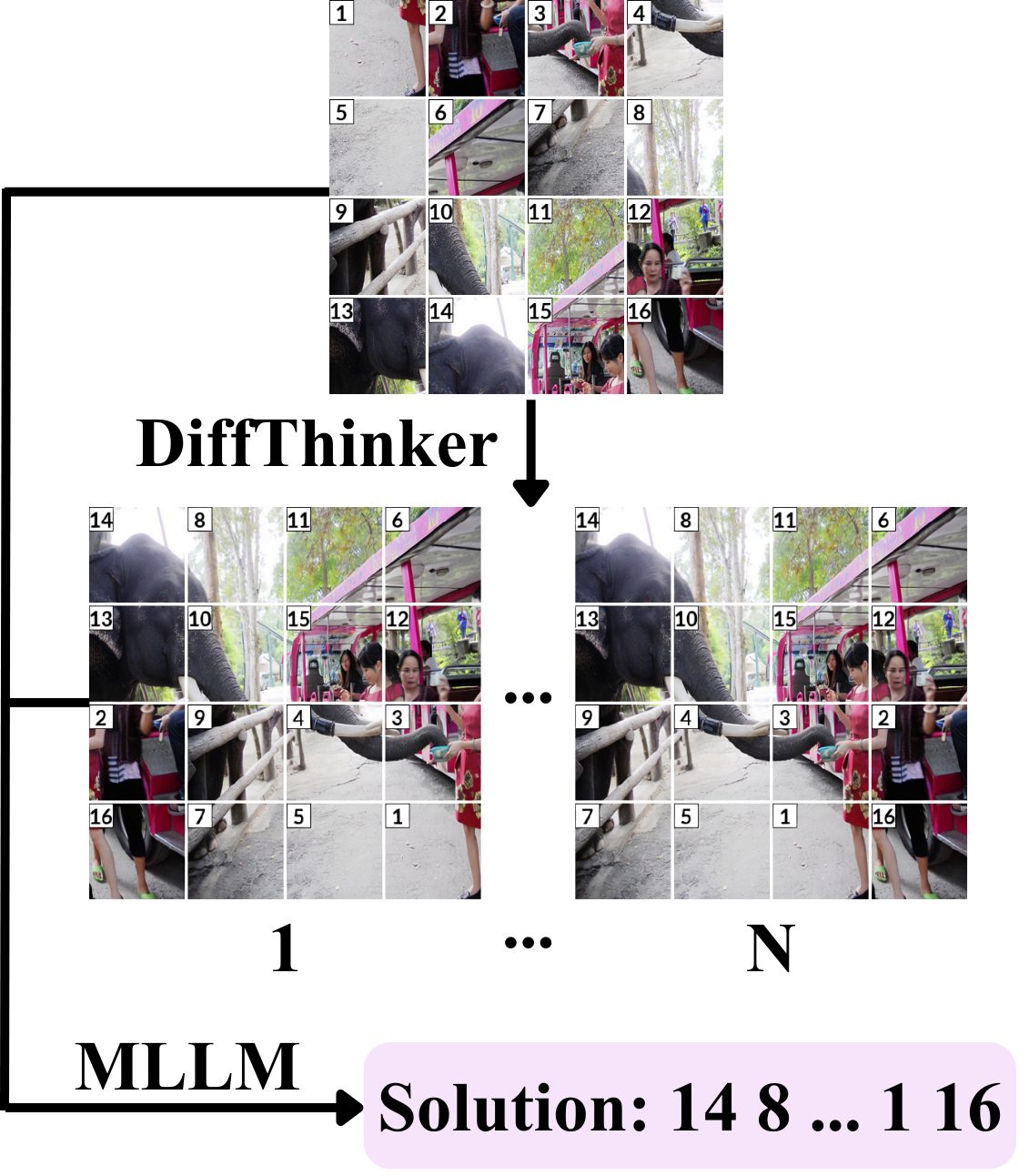}
        
        \vspace{4pt}
        \centerline{\footnotesize (a) Collaborative Pipeline}
    \end{minipage}
    \hfill
    \begin{minipage}{0.6\columnwidth}
        \centering
        \includegraphics[width=\textwidth]{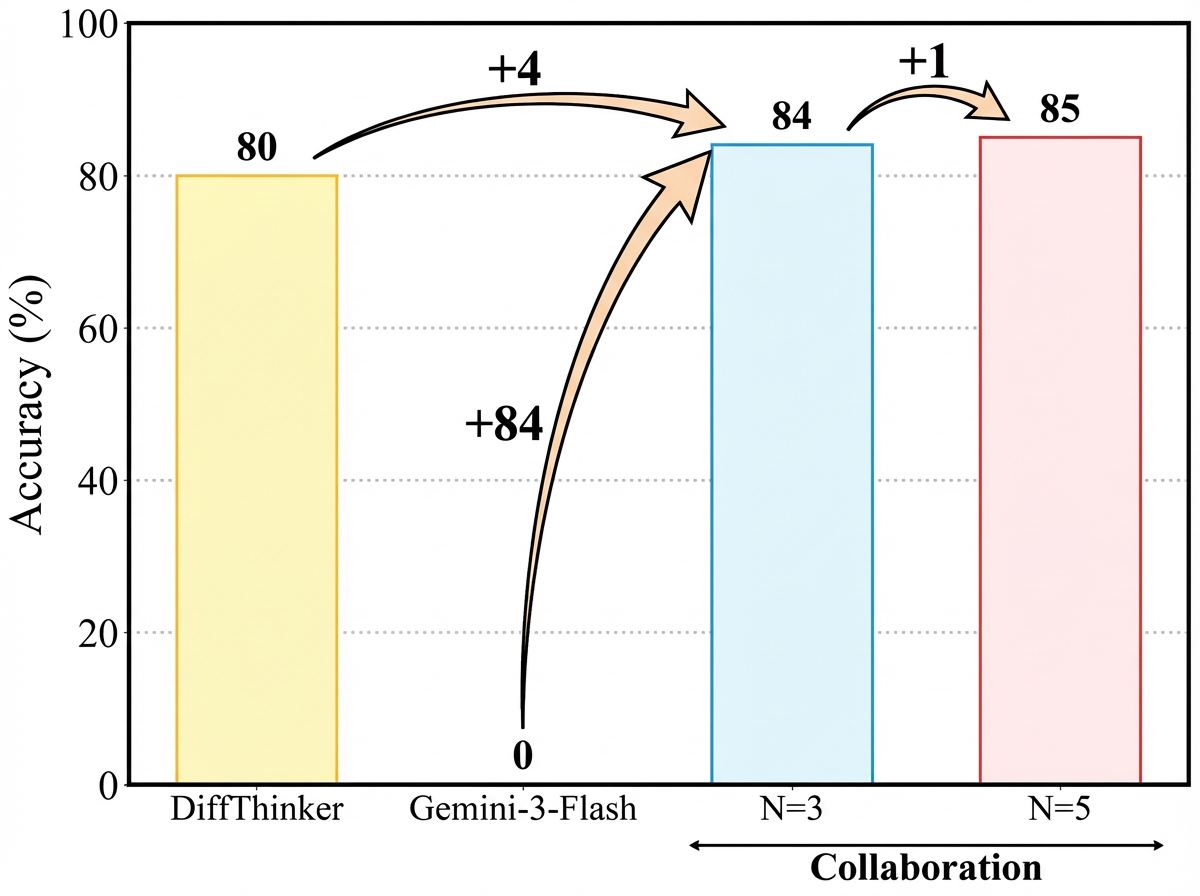}
        \centerline{\footnotesize (b) Accuracy on Jigsaw level-4}
    \end{minipage}
    \caption{\textbf{DiffThinker as a collaborative partner}. (a) The partnership framework where DiffThinker generates $N$ candidates for MLLM verification. (b) Performance on Jigsaw level-4, demonstrating that collaboration surpasses individual models and accuracy further scales with the number of candidates $N$.
}
    \label{fig:collaboration}
\end{figure}

\textbf{DiffThinker as a Collaborative Partner.} 
Beyond a direct comparison, we explore the synergy between DiffThinker and MLLMs in solving complex tasks. 
As illustrated in Figure~\ref{fig:collaboration}, DiffThinker first produces multiple candidate solution images, which the MLLM then evaluates against the original problem constraints to make a final decision. 

We benchmark this collaborative approach on Jigsaw level-4, which demands both spatial reasoning and rigorous verification. Results demonstrate that this partnership achieves superior accuracy, outperforming either model in isolation. 
Specifically, DiffThinker compensates for the MLLM’s limited visual imagination in spatial reasoning, while the MLLM leverages its reflective capabilities to filter potential errors in the generated candidates. 
This synergy reveals that DiffThinker can serve as a powerful visual reasoning backend to augment the cognitive breadth of MLLMs.

\begin{figure}[t]
\centering
\includegraphics[width=1.0\linewidth]{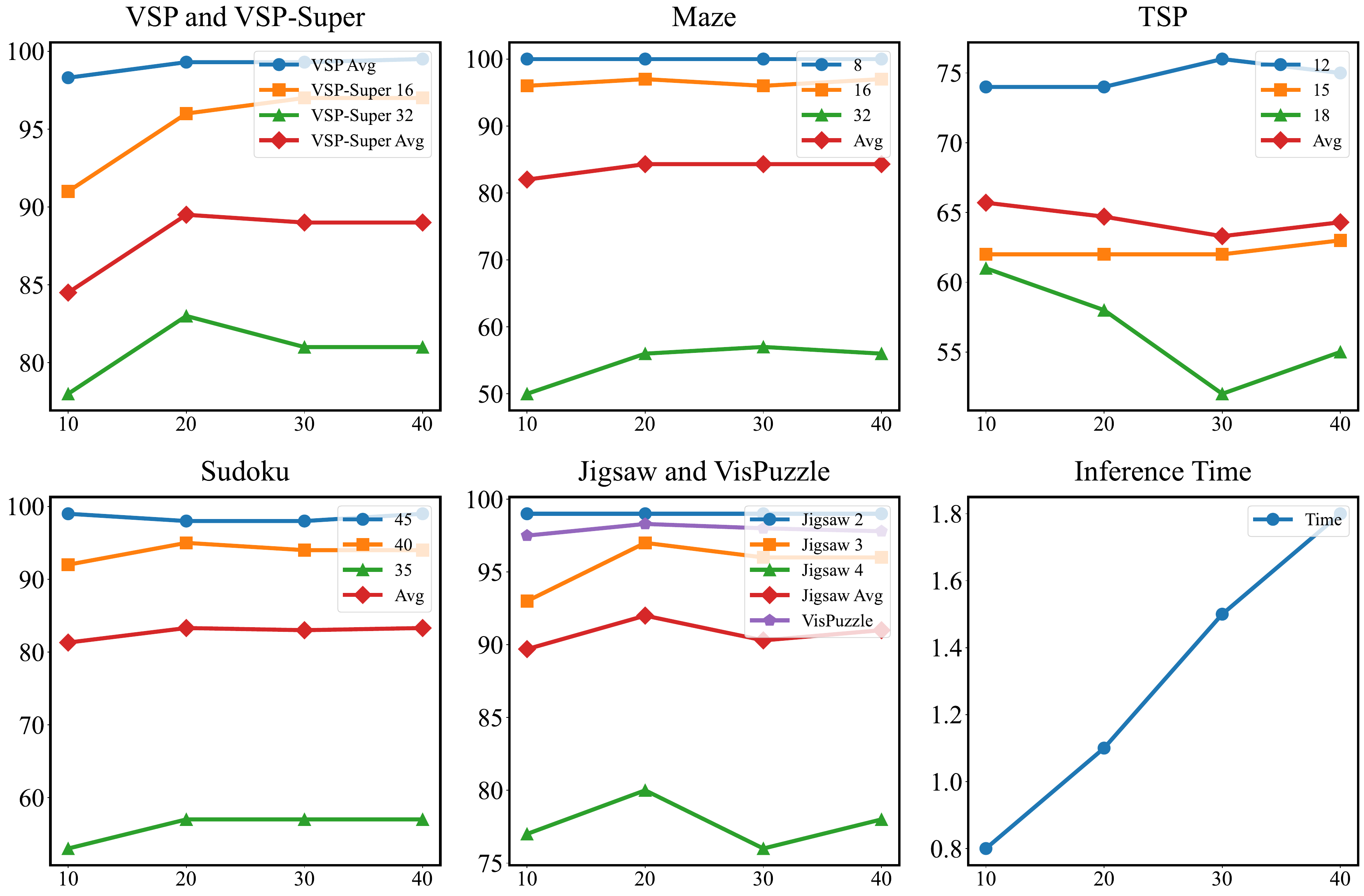}
\caption{\textbf{Trade-off between accuracy and inference time across varying inference steps.} The horizontal axis denotes the number of inference steps, while the vertical axis denotes accuracy or inference time. An optimal balance between reasoning performance and computational cost is achieved at approximately 20 steps.}

\label{fig:step_sensitivity}
\end{figure}

\subsection{Ablation Studies}

\textbf{Ablation on Inference Steps.}
We first investigate the trade-off between accuracy and inference time, as illustrated in Figure~\ref{fig:step_sensitivity}. DiffThinker demonstrates remarkable robustness, maintaining high performance even with as few as 10 inference steps. Increasing the step count to 20 yields a noticeable performance boost, identifying an optimal balance between solution quality and computational efficiency. Beyond 20 steps, the accuracy plateaus with only marginal fluctuations, suggesting that the underlying reasoning manifold is effectively captured early in the generative process. Based on these observations, we adopt 20 inference steps as our default configuration for evaluations to ensure superior performance with minimal inference overhead.

\begin{figure}[t]
    \centering
    \begin{subfigure}[b]{1.0\linewidth}
        \centering
        \includegraphics[width=1.0\linewidth]{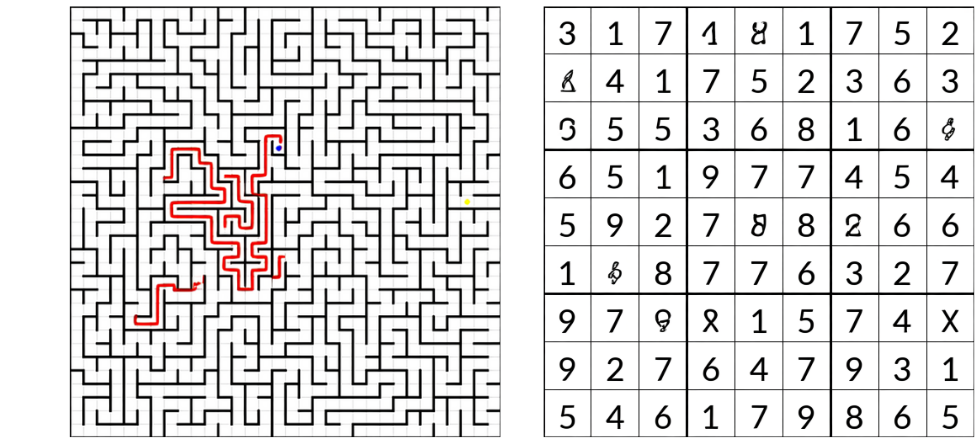}
        \caption{Qualitative analysis with 100 training samples.}
        \label{fig:low_data_samples}
    \end{subfigure}
    
    \begin{subfigure}[b]{1.0\linewidth}
        \centering
        \includegraphics[width=1.0\linewidth]{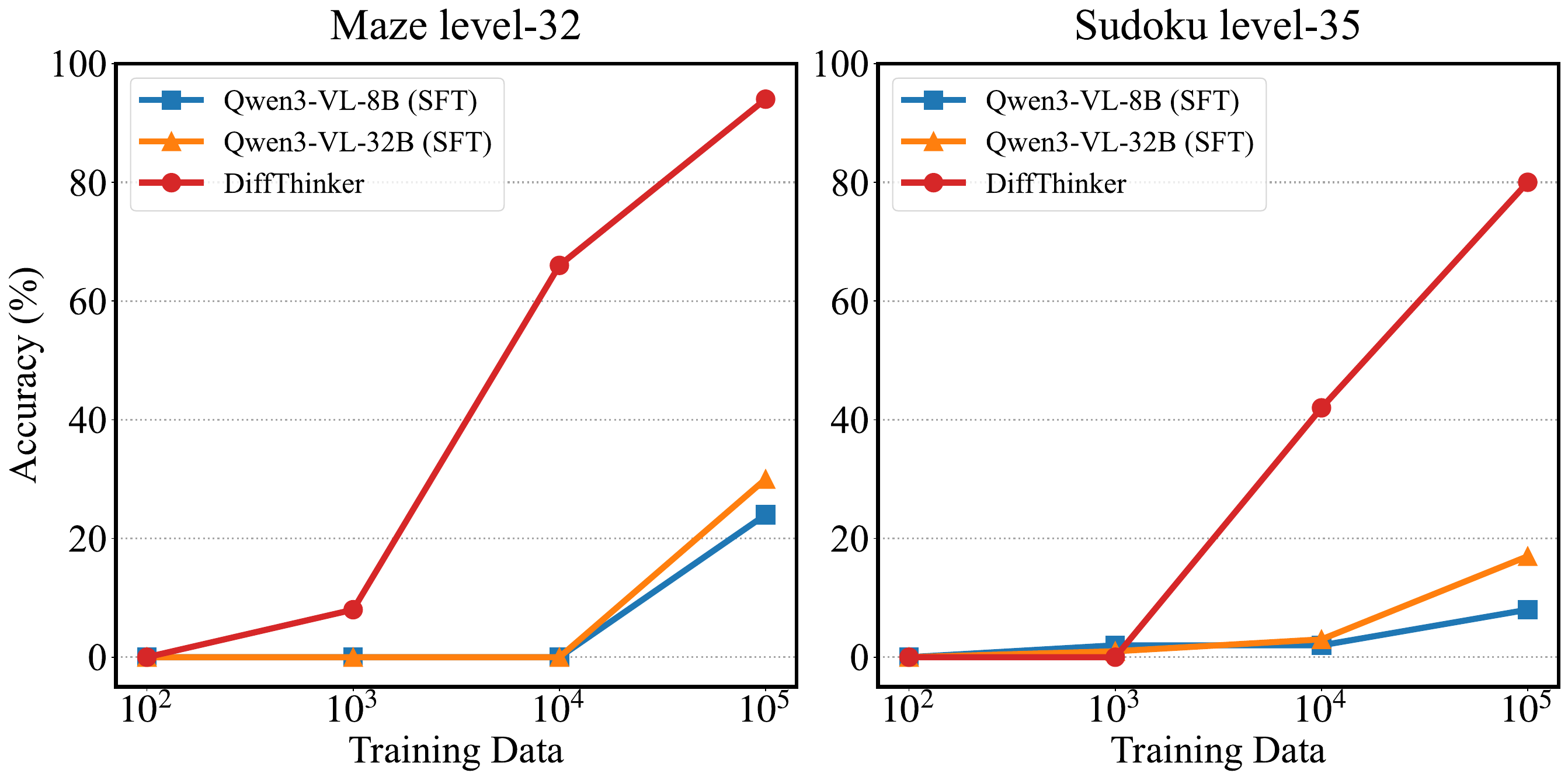}
        \caption{Quantitative analysis with increasing training samples.}
        \label{fig:scaling_curves}
    \end{subfigure}

    \caption{\textbf{Ablation on Training Data Scale.} (a) Qualitative analysis shows that with limited data, DiffThinker focuses on mastering rendering syntax. (b) Quantitative results on Maze level-32 and Sudoku level-35 demonstrate that DiffThinker scales consistently with data expansion.}
    \label{fig:comprehensive_scaling}
\end{figure}

\textbf{Ablation on Training Data Scale.}
We evaluate the influence of training data size on DiffThinker's performance using our most complex tasks, Maze level-32 and Sudoku level-35. We first qualitatively analyze the model's behavior under low-data regimes. As illustrated in Figure~\ref{fig:comprehensive_scaling}(a), due to the limited zero-shot reasoning capacity of the base model, DiffThinker initially focuses on mastering task-specific rendering syntax, such as grid alignment and trajectory continuity. 
As the training volume increases, DiffThinker transitions from superficial visual imitation to deep structural reasoning. Quantitative results in Figure~\ref{fig:comprehensive_scaling}(b) show that DiffThinker continues to benefit from data expansion, maintaining a consistent upward trajectory. With $10^5$ samples, the model effectively internalizes underlying causal structures, achieving over 90\% accuracy on Maze level-32, while the performance of MLLMs remain significantly limited despite the increased data. Based on these observations, we utilize a total of 30,000 samples across all difficulty levels for each task in our main experiments to achieve an optimal balance between performance and efficiency.

\begin{figure}[t]
    \centering
    \begin{subfigure}[b]{1.0\linewidth}
        \centering
        \includegraphics[width=1.0\linewidth]{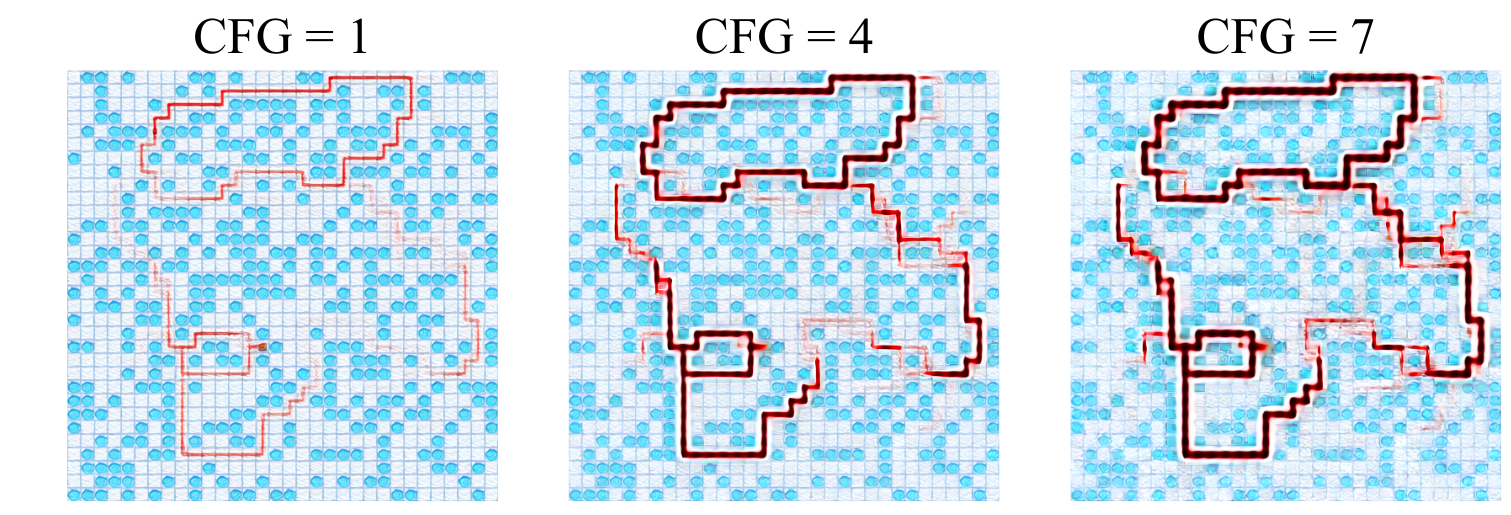}
        \caption{Qualitative results of the predicted sample $\hat{x}_0$ at step 1.}
        \label{fig:cfg_qualitative}
    \end{subfigure}

    \begin{subfigure}[b]{1.0\linewidth}
        \centering
        \includegraphics[width=1.0\linewidth]{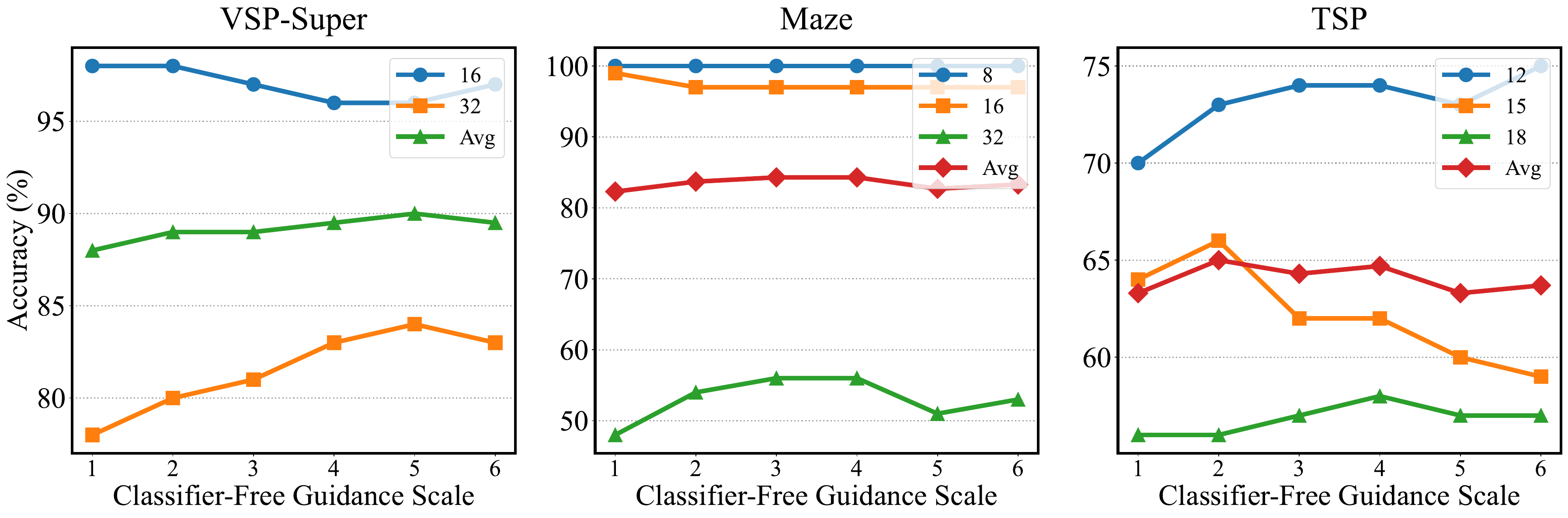}
        \caption{Quantitative results of accuracy relative to CFG scales.}
        \label{fig:cfg_quantitative}
    \end{subfigure}

    \caption{\textbf{Ablation analysis of Classifier Free Guidance scales.} (a) Impact of CFG scales on path clarity. (b) Accuracy trends across tasks confirming $w=4$ as the peak performance point for balancing logic and fidelity.}
    \label{fig:cfg_analysis}
\end{figure}

\textbf{Ablation on Classifier-Free Guidance Scale.} 
We investigate the impact of Classifier-Free Guidance (CFG)~\cite{ho2022classifier} on the reasoning capabilities of DiffThinker. 
As a core mechanism in diffusion models, CFG regulates the trade-off between conditional adherence and sample fidelity. 
The guided velocity field $\hat{v}_\theta$ is defined as:
\begin{equation}
    \hat{v}_\theta(x_t, t, h) = v_\theta(x_t, t, \emptyset) + w (v_\theta(x_t, t, h) - v_\theta(x_t, t, \emptyset))
\end{equation}
where $v_\theta(x_t, t, h)$ and $v_\theta(x_t, t, \emptyset)$ represent the conditional and unconditional velocity predictions, respectively, and $w$ denotes the CFG scale. 
We begin with a qualitative assessment of different CFG scales. Figure~\ref{fig:cfg_analysis}(a) visualizes the predicted original sample $\hat{x}_0$ at the first step across varying scales. At $w=1$, the insufficient conditioning produces faint and tentative trajectories, lacking the deterministic confidence for logical precision. 
Conversely, $w=7$ triggers numerical over-saturation and visual artifacts, leading to distorted textures that severely degrade generative fidelity.  
Between these extremes, $w=4$ effectively acts as a logic amplifier, generating bold and precise paths that perfectly align with task constraints. 

Quantitatively, Figure~\ref{fig:cfg_analysis}(b) demonstrates that reasoning performance is robust across various guidance scales, with accuracy peaking at $w=4$ across the majority of levels.  
Consequently, we adopt $w=4$ as the default configuration for all experiments to ensure an optimal balance between logical precision and generative fidelity.

\subsection{Image Generation vs. Video Generation.} 
Video generation offers unique advantages for multimodal reasoning by explicitly modeling temporal coherence and the continuous evolution of state transitions. Its capacity to represent reasoning trajectories as a fluid sequence could naturally facilitate the resolution of complex planning tasks. Motivated by these potential benefits, we explore the feasibility of video-based reasoning and conduct a direct comparison with our image-based approach. Our video-based baseline, denoted as DiffThinker-Video, is implemented upon Wan2.2-TI2V-5B~\cite{wan}, a leading open source video foundation model. Due to the relatively weaker reasoning proficiency observed in current video generation models, we perform training and evaluation on Maze level-8, a relatively simple task that is also well-suited for video-based reasoning. To ensure a fair comparison, we train both models on identical datasets for varying numbers of epochs and report training duration and corresponding accuracy.

\begin{figure}[t]
    \centering
    \begin{minipage}{0.45\columnwidth}
        \centering
        \includegraphics[width=\linewidth]{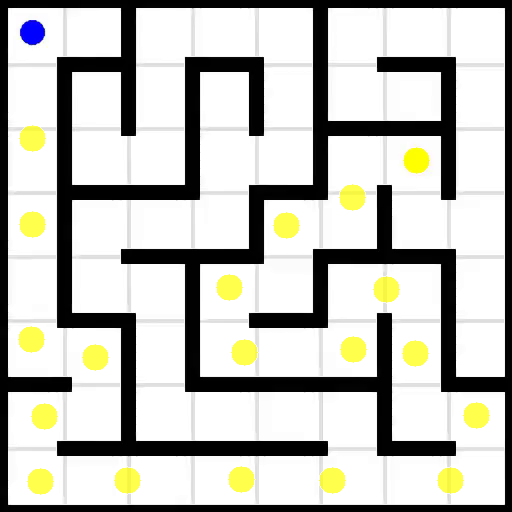}
        \caption{\textbf{Visual Trajectory of DiffThinker-Video.} Visualized through accumulation of uniformly sampled frames.}
        \label{fig:video_demo}
    \end{minipage}
    \hfill
    \begin{minipage}{0.52\columnwidth}
        \centering
        {\includegraphics[width=\linewidth]{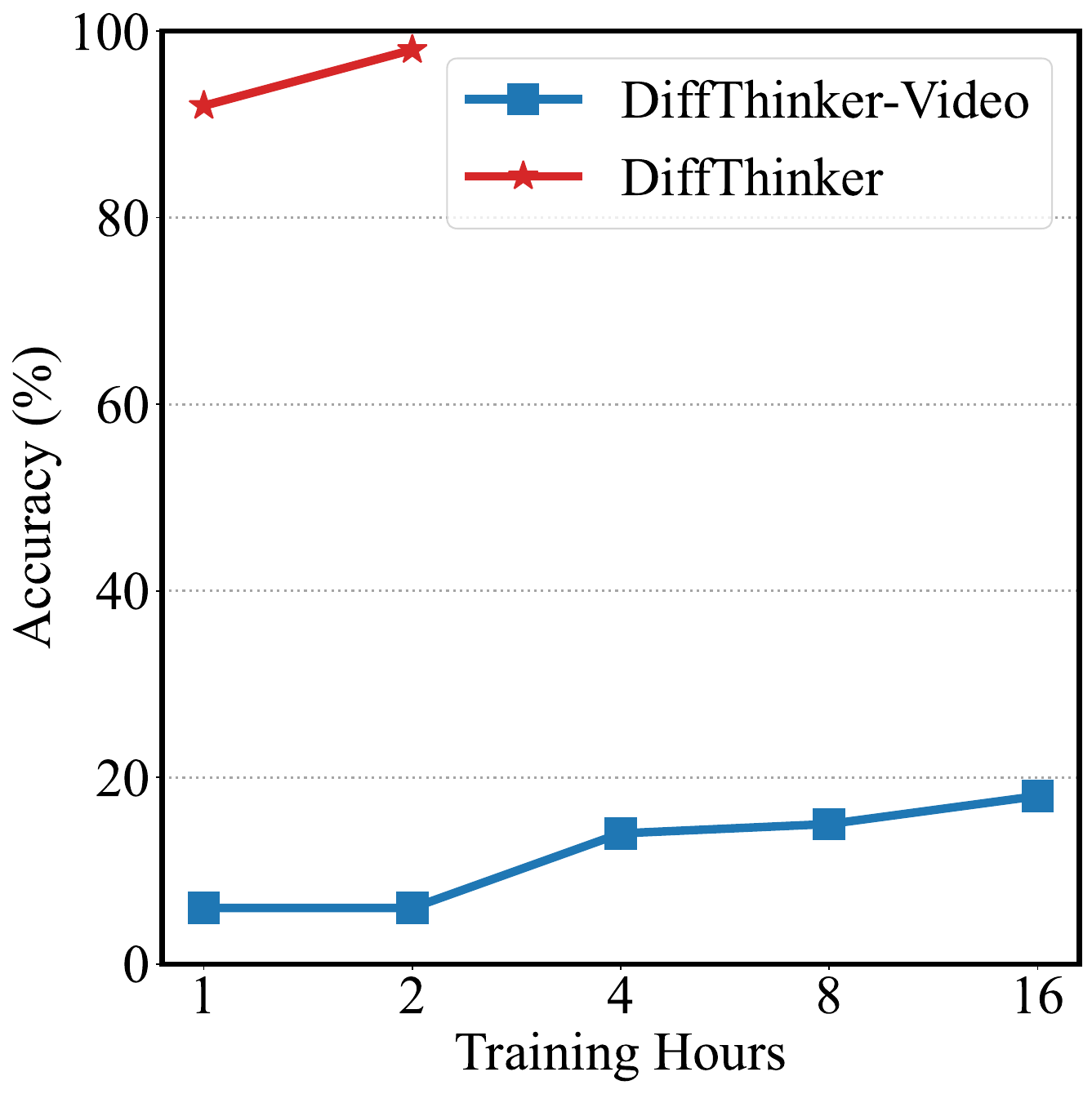}}
        \vspace{-2mm} 
        \caption{\textbf{Performance comparison between two paradigms.}}
        \label{fig:video_gen}
    \end{minipage}
\end{figure}

Qualitatively, Figure~\ref{fig:video_demo} demonstrates that DiffThinker-Video possesses inherent reasoning capabilities; it resolves the maze problem by generating a video where a yellow ball progressively navigates the paths toward the target. Quantitatively, however, Figure~\ref{fig:video_gen} reveals that it yields lower accuracy with higher training overhead than DiffThinker. 
Furthermore, despite its smaller parameter count, DiffThinker-Video requires 2.0s per inference, nearly doubling the 1.1s latency of DiffThinker. 
These results highlight the prohibitive computational costs of video generation, underscoring the need for more efficient video models to advance generative multimodal reasoning.

\section{Conclusion}
In this paper, we introduce DiffThinker and establish Generative Multimodal Reasoning as a novel paradigm for complex vision-centric tasks. By leveraging diffusion models, we reformulate multimodal reasoning from a traditional text-centric symbolic mapping into a native generative image-to-image task, enabling models to perform reasoning within the visual space with superior logical consistency and spatial precision. Extensive experiments across four domains (sequential planning, combinatorial optimization, constraint satisfaction, and spatial configuration) demonstrate that DiffThinker significantly outperforms state-of-the-art MLLMs. Our systematic analysis further reveals the intrinsic advantages of this paradigm, including its efficiency, controllability, and native parallelism, while showcasing its potential as a collaborative backend to augment the cognitive breadth of MLLMs. We hope DiffThinker will inspire further exploration into Generative Multimodal Reasoning to unlock the full potential of multimodal intelligent agents.

{
\bibliography{main}
\bibliographystyle{icml2026}
}

\clearpage
\newpage
\appendix
\onecolumn

\section{Implementation Details}

\subsection{Training Details}
\label{sec:training_details}

\subsubsection{Data Preparation.} 
The datasets utilized for training and evaluation are detailed in Table~\ref{tab:data_specs}. Following previous research~\cite{wang2025jigsaw,wu2025visual}, we utilize the COCO~\cite{lin2014microsoft} dataset to synthesize samples for both the training and testing of jigsaw puzzles. Specifically, we instantiate five independent models, each specialized for one of the five task categories, and subsequently evaluate them on their respective test benchmarks. All training datasets undergo thorough deduplication. Both DiffThinker and the baseline MLLMs are trained on identical data distributions to ensure an equitable comparison.

\subsubsection{Hyperparameter Configuration.} 
We summarize the key training configurations and hyperparameters for Flow Matching, SFT, and GRPO in Table~\ref{tab:training_hyperparams}. In accordance with common practices, we employ Low-Rank Adaptation (LoRA)~\cite{lora} for both the fine-tuning of Qwen-Image-Edit and the SFT of Qwen3-VL. For GRPO, considering the substantial computational overhead associated with reinforcement learning, we limit the training to a single epoch and utilize a reduced rollout number to maintain a manageable training budget while ensuring comparability across different experimental settings.

\begin{table*}[t]
\centering
\caption{\textbf{Detailed statistics for training and testing datasets across five task categories.}}
\label{tab:data_specs}
\small
\begin{tabular}{llcc}
\toprule
\textbf{Task Category} & \textbf{Difficulty Levels} & \textbf{Training Samples} & \textbf{Test Samples} \\
\midrule
 & 3, 4, 5, 6 (Grid size) & 500, 1,000, 2,500, 6,000 & 100 per level * \\
\textbf{VSP \& VSP-Super}& 7, 8 (Grid size) & -- & 100 per level $^{\dagger}$ \\
& 16, 32 (Grid size) & 10,000, 10,000 & 100 per level  \\
\midrule
\textbf{Maze} & 8, 16, 32 (Grid size) & 10,000  & 100 per level  \\
\midrule
 & 12,  15 (City count) & 5000  & 100 per level  \\
\textbf{TSP} & 13, 14, 16, 17 (City count) & 5000  & -- \\
& 18 (City count) & -- & 100 per level$^{\dagger}$ \\
\midrule
\textbf{Sudoku} & 30 (Number of given clues) & 7,500  & -- \\
 & 35, 40, 45 (Number of given clues) & 7,500  & 100 per level\\
\midrule
 & $1{\times}2, 1{\times}3, 2{\times}1, 3{\times}1$ (Patch layout) & 4000 per level & -- \\
\textbf{Jigsaw \& VisPuzzle}&$2{\times}2, 3{\times}3, 4{\times}4$ (Patch layout) & 4,000, 5,000, 5,000  &  100 per level \\
& VisPuzzle & -- & 400*$^{\dagger}$ \\
\bottomrule
\end{tabular}
\begin{flushleft}
\hspace{2em} \footnotesize * denotes tasks utilizing official benchmarks from prior works. \quad $^{\dagger}$ denotes out-of-distribution testing scenarios.
\end{flushleft}
\end{table*}

\begin{table*}[t]
\centering
\caption{\textbf{Hyperparameter settings for different training  paradigms.}}
\label{tab:training_hyperparams}
\small
\begin{tabular}{lccc}
\toprule
& \textbf{Flow Matching} & \textbf{SFT} & \textbf{GRPO} \\
\midrule
Framework & DiffSynth-Studio~\cite{diff} & SWIFT~\cite{swift} & verl~\cite{verl} \\
Epochs & 5 & 5 & 1 \\
Learning Rate & $1 \times 10^{-4}$ & $1 \times 10^{-4}$ & $1 \times 10^{-6}$ \\
LoRA Rank & 32 & 32 & -- \\
Batch Size & 8 & 32 & 128 (8B) / 64 (32B) \\
Rollout Size ($n$) & -- & -- & 4 \\
KL Coefficient & -- & -- & $1 \times 10^{-2}$ \\
\bottomrule
\end{tabular}
\end{table*}

\subsubsection{Reward Functions for GRPO}

Due to the limited zero-shot accuracy of the Qwen3-VL baselines on complex reasoning tasks, employing a strict binary reward based on exact matching results in extremely sparse signals, which significantly hinders the policy optimization process. Therefore, we design task-specific partial reward functions for each domain as follows:

\textbf{Sequential Planning (VSP, VSP-Super, and Maze).} For navigation tasks, we utilize a prefix matching reward. The reward evaluates the longest continuous sequence of correct actions from the starting point to ensure the model learns the correct trajectory incrementally. Given a predicted action sequence $P = (p_1, p_2, \dots, p_m)$ and the ground truth $G = (g_1, g_2, \dots, g_n)$, the reward is defined as:
\begin{equation}
R_{\text{plan}} = \frac{\max \{ k \mid \forall i \leq k, p_i = g_i \text{ and } k \leq \min(m, n) \}}{n}.
\end{equation}

\textbf{Combinatorial Optimization (TSP).} For the Traveling Salesperson Problem, the reward is designed to account for both coordinate set consistency and path length precision. Let $\mathcal{S}_p$ and $\mathcal{S}_g$ denote the sets of coordinates in the predicted and ground truth paths, and $L(\cdot)$ represent the total Euclidean distance of a trajectory. The reward is formulated as follows:
\begin{equation}
R_{\text{TSP}} = 
\begin{cases} 
0.5 \left( 1 + \mathbb{I}(|L(P) - L(G)| < \epsilon) \right) & \text{if } \mathcal{S}_p = \mathcal{S}_g \\
0 & \text{otherwise}
\end{cases},
\end{equation}
where $\epsilon = 1 \times 10^{-4}$ serves as the tolerance for floating point comparisons. This tiered structure ensures that the model is first rewarded for identifying all required cities before optimizing the visitation order to match the ground truth distance.

\textbf{Constraint Satisfaction (Sudoku).} The Sudoku reward is based on the element wise accuracy of the completed grid. We first normalize the model output by extracting all numeric digits to form the predicted sequence $P$. If the length of the predicted sequence $|P|$ matches the standard 81 digits required for a $9 \times 9$ grid, the reward is calculated as the proportion of correctly filled cells. The reward function is defined as:
\begin{equation}
R_{\text{Sudoku}} = 
\begin{cases} 
\frac{1}{81} \sum_{i=1}^{81} \mathbb{I}(p_i = g_i) & \text{if } |P| = 81 \\
0 & \text{otherwise}
\end{cases},
\end{equation}
where $\mathbb{I}(\cdot)$ denotes the indicator function and $g_i$ represents the ground truth value for the $i$-th cell. This objective encourages the model to respect both the structural integrity of the grid and the specific numerical constraints of the puzzle.

\textbf{Spatial Configuration (Jigsaw and VisPuzzle).} For Jigsaw tasks, the reward measures the positional accuracy of the restored image patches. After normalizing the predicted sequence $P$ and the ground truth sequence $G$ by removing extraneous whitespace, we evaluate the element wise matching rate. If the length of the predicted sequence $|P|$ equals the total number of patches $n$, the reward is defined as the proportion of patches assigned to their correct absolute positions:
\begin{equation}
R_{\text{Jigsaw}} = 
\begin{cases} 
\frac{1}{n} \sum_{i=1}^{n} \mathbb{I}(p_i = g_i) & \text{if } |P| = n \\
0 & \text{otherwise}
\end{cases}.
\end{equation}

\subsection{Prompt}

\begin{figure*}[ht]
    \centering
    \includegraphics[width=0.98\linewidth]{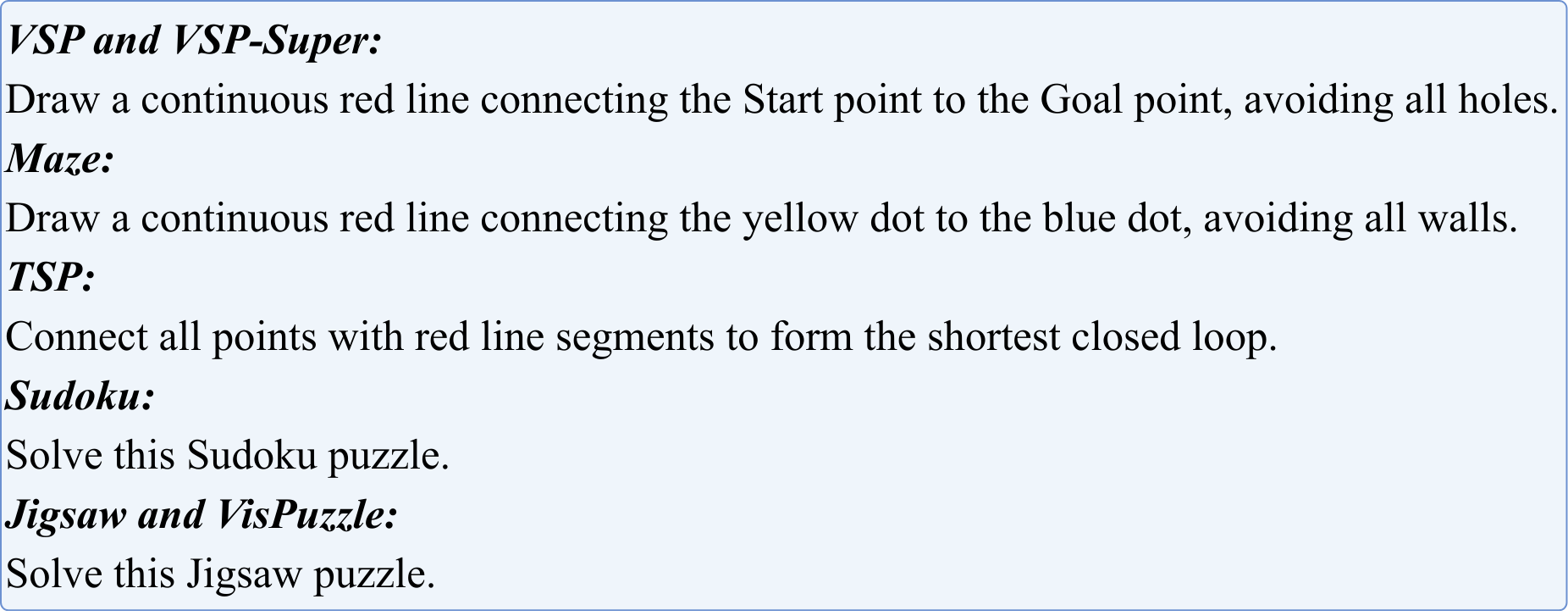} \\ 
    \caption{\textbf{Prompt templates for DiffThinker.}}
    \label{fig:prompts_p1}
\end{figure*}

\begin{figure*}[ht]
    \centering
    \includegraphics[width=0.98\linewidth]{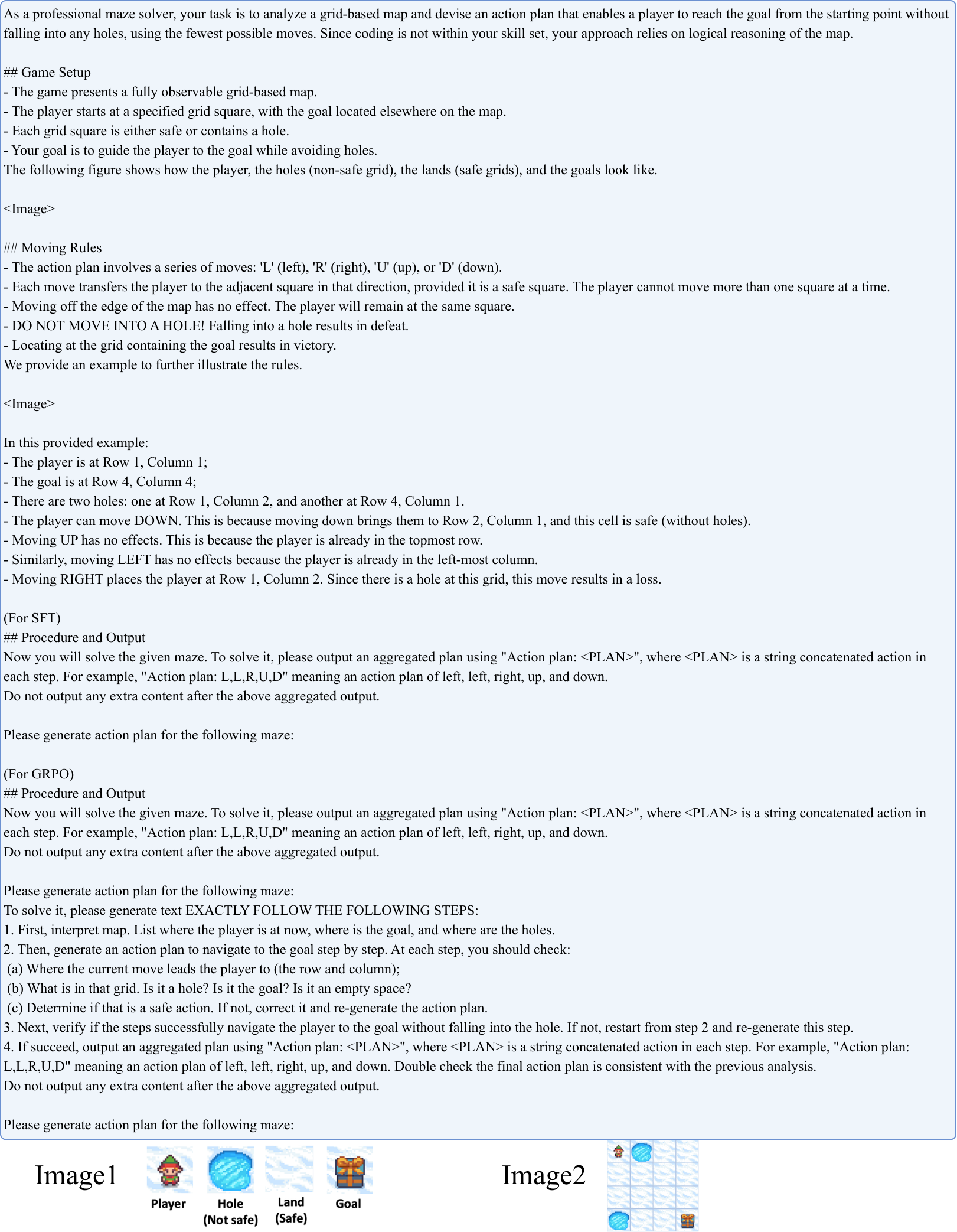} \\ 
    \caption{\textbf{Prompt templates of VSP and VSP-Super for MLLMs.}}
    \label{fig:prompts_p2}
\end{figure*}

\begin{figure*}[ht]
    \centering
    \includegraphics[width=0.98\linewidth]{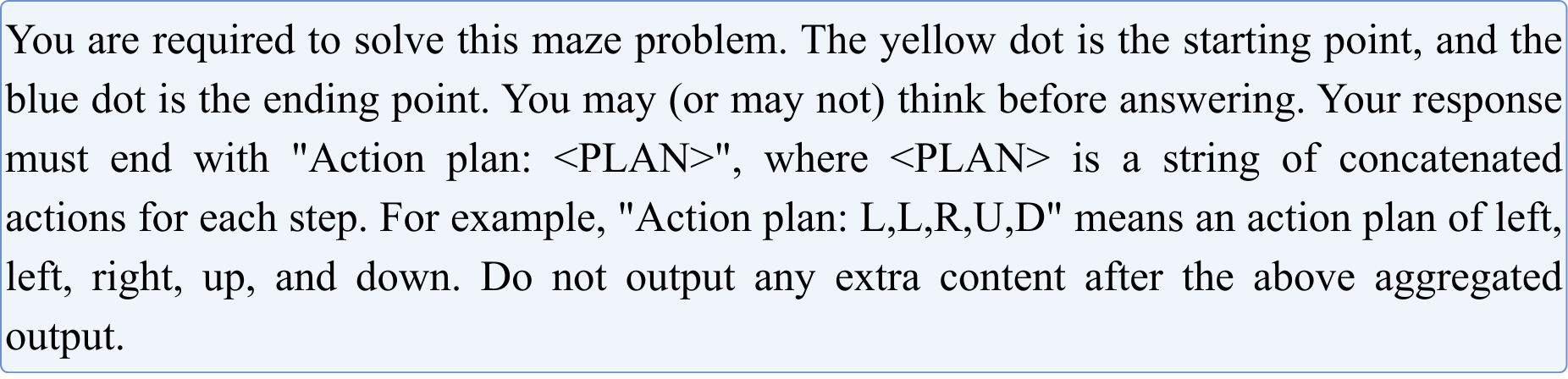} \\ 
    \caption{\textbf{Prompt templates of Maze for MLLMs.}}
    \label{fig:prompts_p3}
\end{figure*}

\begin{figure*}[ht]
    \centering
    \includegraphics[width=0.98\linewidth]{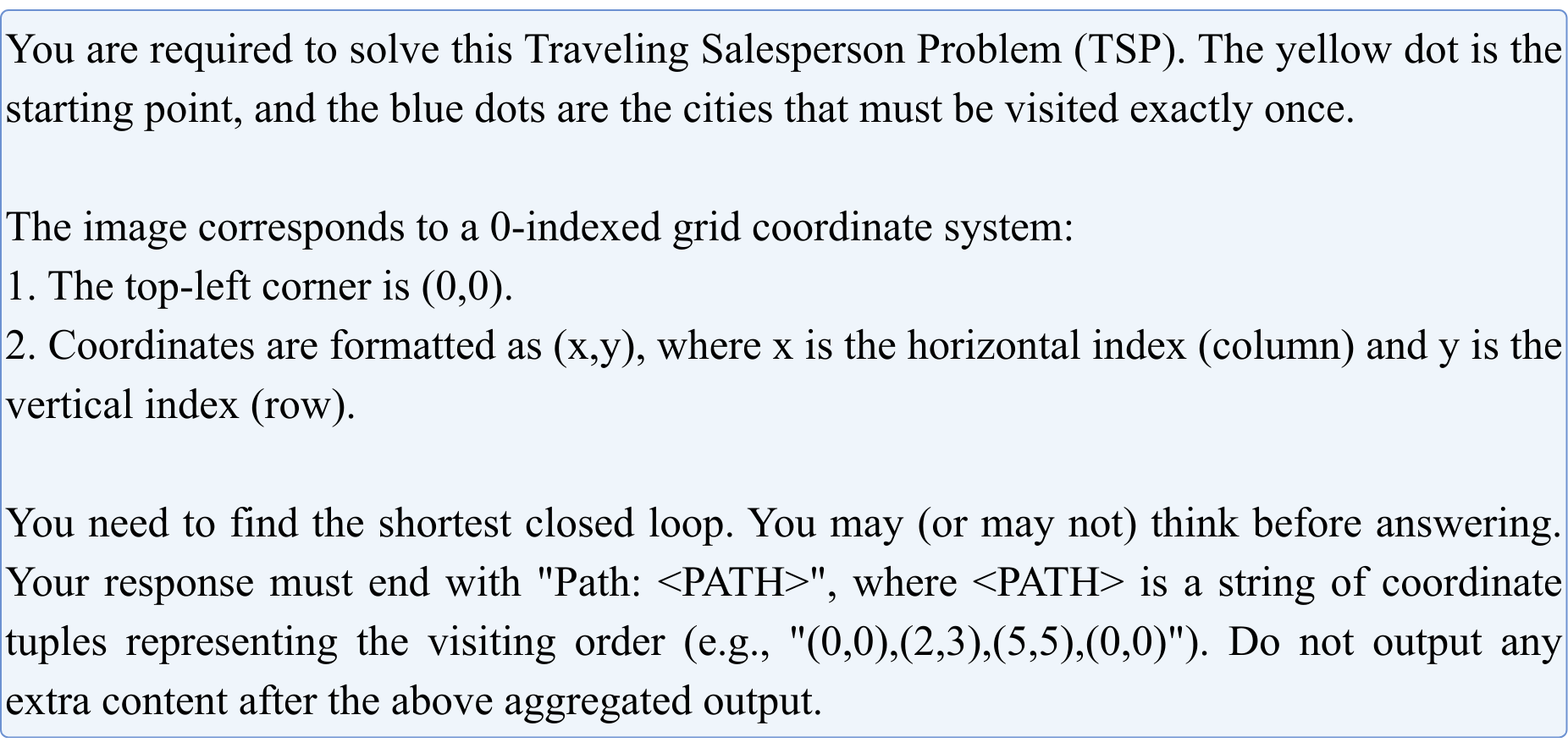} \\ 
    \caption{\textbf{Prompt templates of TSP for MLLMs.}}
    \label{fig:prompts_p4}
\end{figure*}

\begin{figure*}[ht]
    \centering
    \includegraphics[width=0.98\linewidth]{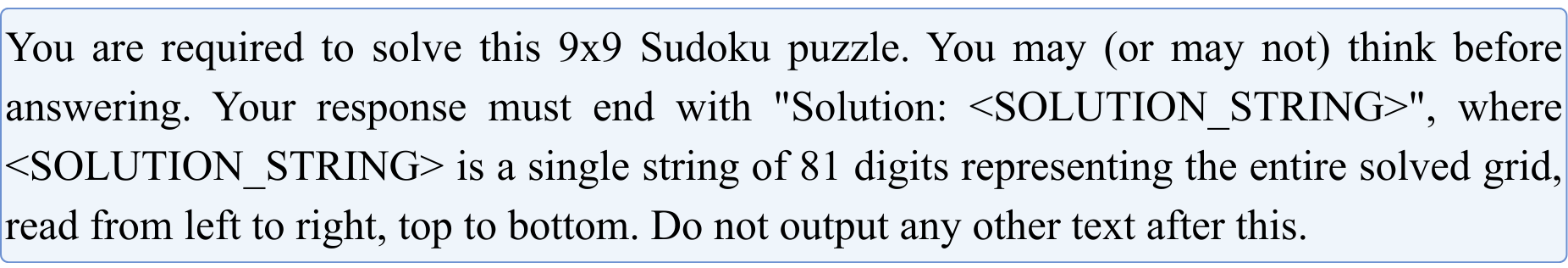} \\ 
    \caption{\textbf{Prompt templates of Sudoku for MLLMs.}}
    \label{fig:prompts_p5}
\end{figure*}

\begin{figure*}[ht]
    \centering
    \includegraphics[width=0.98\linewidth]{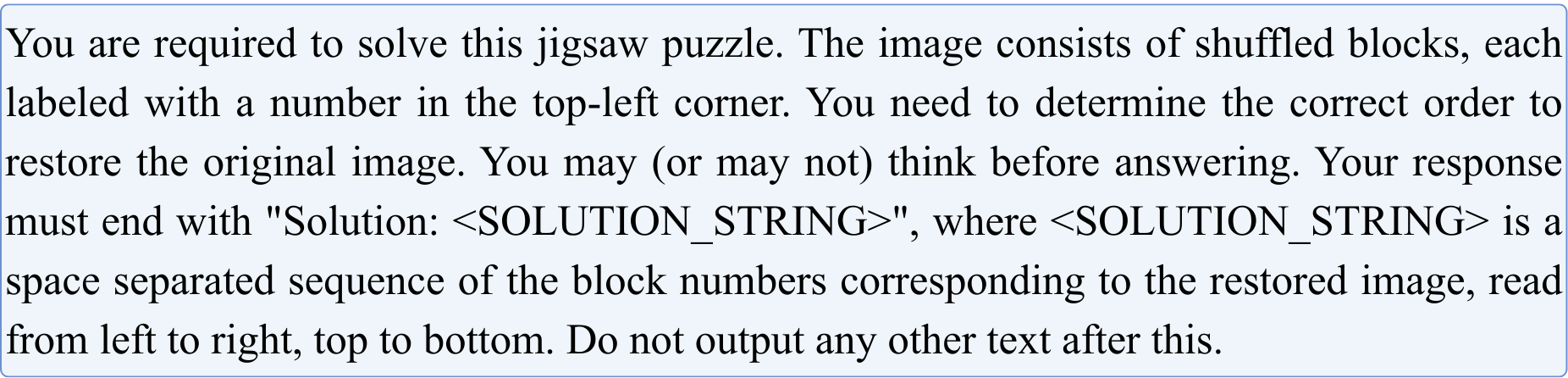} \\ 
    \caption{\textbf{Prompt templates of Jigsaw and VisPuzzle for MLLMs.}}
    \label{fig:prompts_p6}
\end{figure*}

Figures \ref{fig:prompts_p1} through \ref{fig:prompts_p6} provide a comprehensive overview of the prompt templates utilized in our study. For VSP and VSP-Super, we adopt the original prompt~\cite{vsp} designs as specified in the primary literature for the evaluation of Zero-Shot MLLMs. However, for SFT, the prompt structures are specifically adapted as illustrated in Figure \ref{fig:prompts_p2}. This modification is necessitated by the fact that our SFT paradigm does not employ Chain-of-Thought (CoT), requiring a more direct and concise instructional format to ensure consistency with the supervised training objectives.

\section{Limitations and Future Work}

DiffThinker demonstrates state-of-the-art performance in vision-centric reasoning within targeted domains. However, its out-of-distribution (OOD) generalization remains constrained by the limited zero-shot reasoning proficiency of current generative foundation models. Since the reasoning process is directly modeled as a generative task, the model's ability to handle unseen, complex scenarios is heavily tied to the representational depth of its underlying base. Future research should prioritize the development of more robust multimodal generative foundation models specifically optimized for reasoning. Building upon such foundations, we aim to further explore the boundaries of generative multimodal reasoning and enhance its capability to generalize across broader, out-of-distribution tasks.

Furthermore, this work primarily focuses on vision-centric challenges, where DiffThinker significantly surpasses traditional MLLMs. It is important to acknowledge, however, that MLLMs maintain a clear advantage in text-centric domains, such as complex mathematical problems. We do not view these paradigms as mutually exclusive; rather, a promising future direction lies in investigating deeper collaboration and synergy between generative reasoners and MLLMs. By integrating the superior visual precision of DiffThinker with the advanced linguistic and symbolic capabilities of MLLMs, we can extend the scope of multimodal reasoning to a wider spectrum of diverse and demanding tasks.

\section{Qualitative Analysis}

\begin{figure*}[ht]
    \centering
    \includegraphics[width=0.98\linewidth]{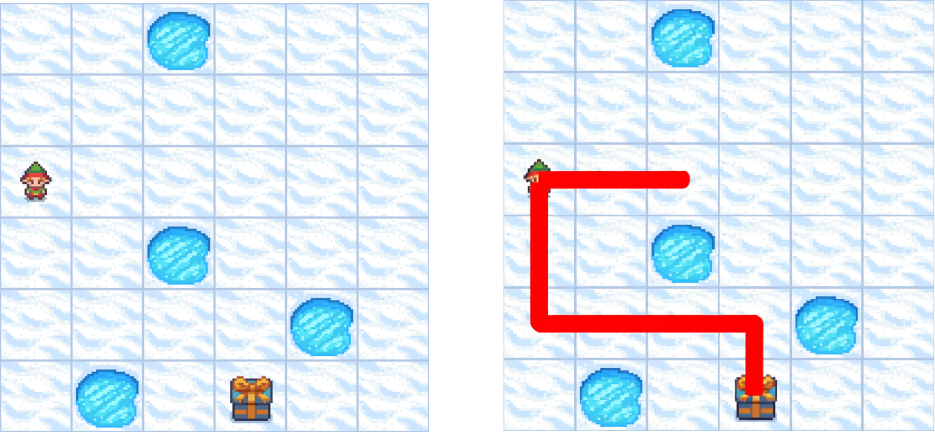} \\ 
    \caption{\textbf{Failure case of DiffThinker on VSP.} In a simple task, DiffThinker performs excessive parallel reasoning but fails to preserve a unique trajectory, ultimately leading to a failure.}
    \label{fig:fail1}
\end{figure*}

\begin{figure*}[ht]
    \centering
    \includegraphics[width=0.98\linewidth]{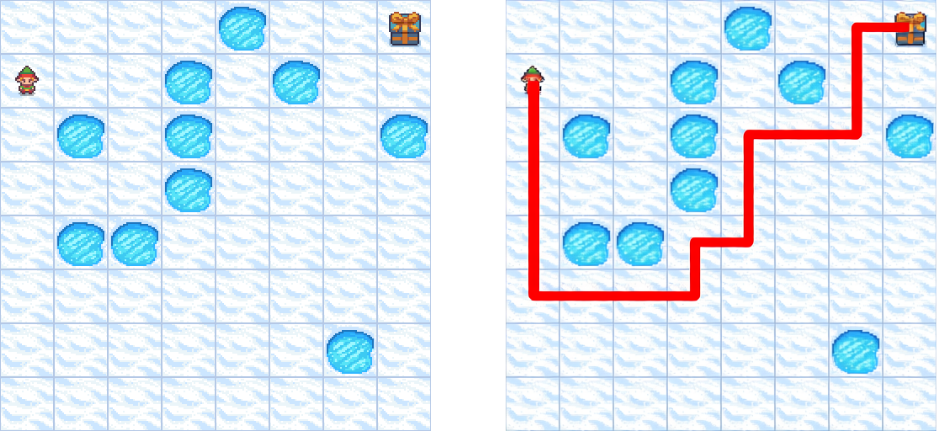} \\ 
    \caption{\textbf{Success case of DiffThinker on VSP.}}
    \label{fig:success1}
\end{figure*}

\begin{figure*}[ht]
    \centering
    \includegraphics[width=0.98\linewidth]{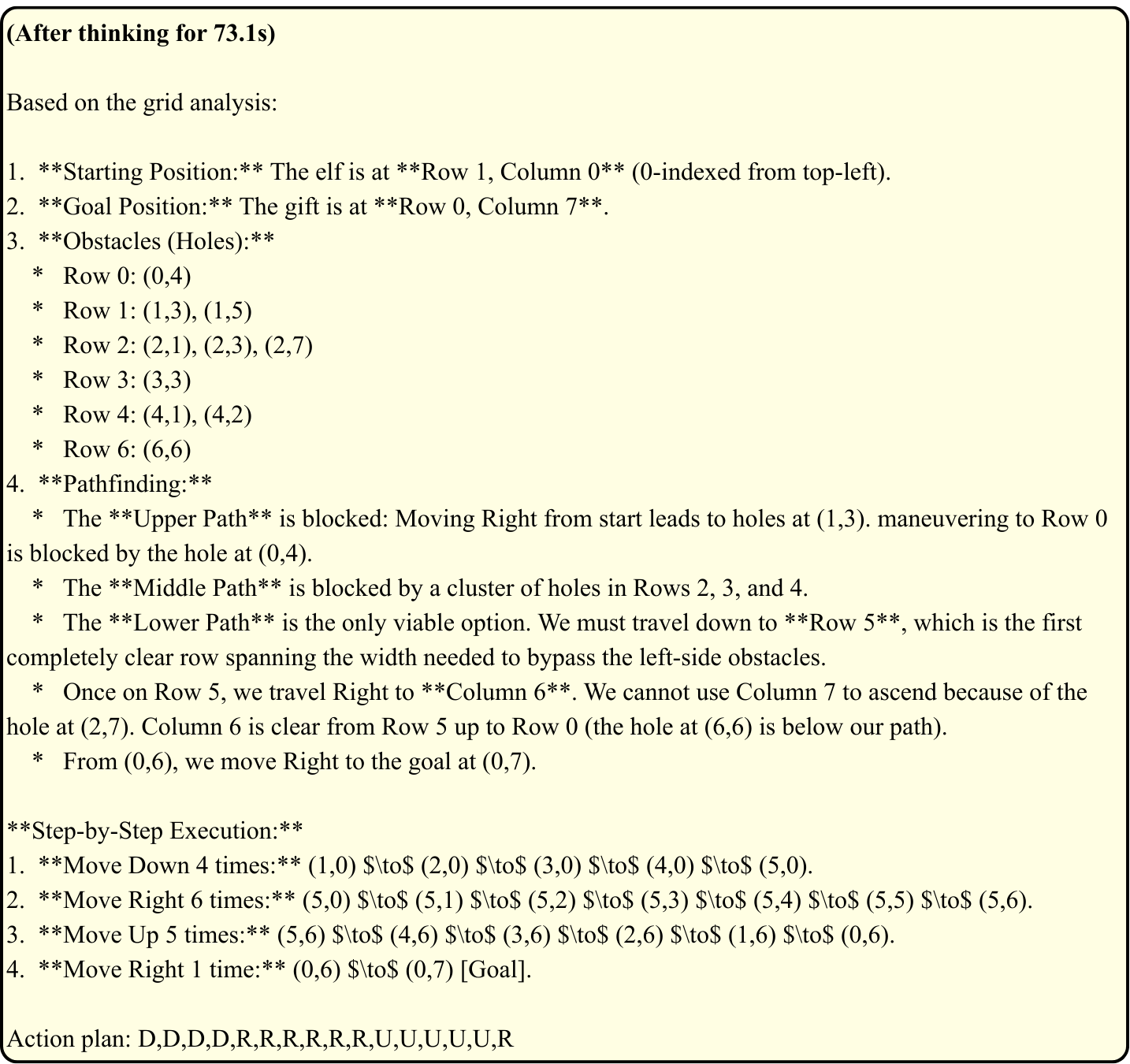} \\ 
    \caption{\textbf{Thinking process of Gemini-3-Pro on VSP.} Gemini-3-Pro successfully provides the correct solution.}
    \label{fig:mllm1}
\end{figure*}

\begin{figure*}[ht]
    \centering
    \includegraphics[width=0.98\linewidth]{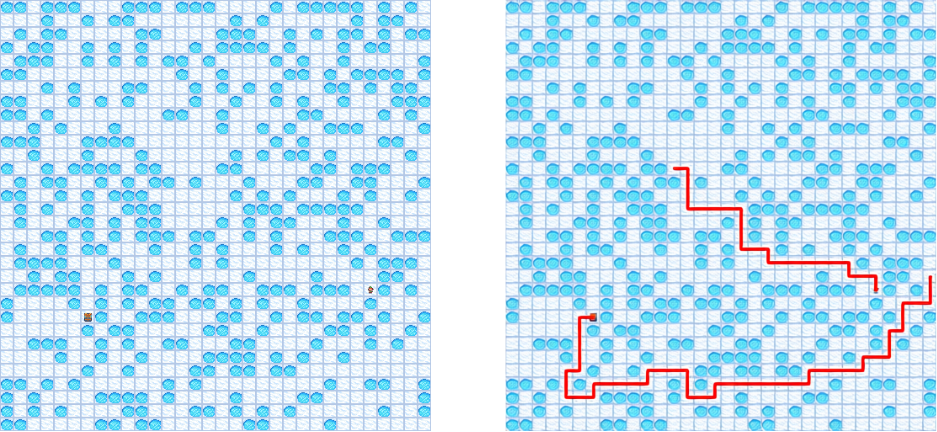} \\ 
    \caption{\textbf{Failure case of DiffThinker on VSP-Super.} In a complex task, DiffThinker identifies a nearly correct trajectory; however, the path is obstructed by a hole, preventing further progress and leading to an ultimate failure.}
    \label{fig:fail2}
\end{figure*}

\begin{figure*}[ht]
    \centering
    \includegraphics[width=0.98\linewidth]{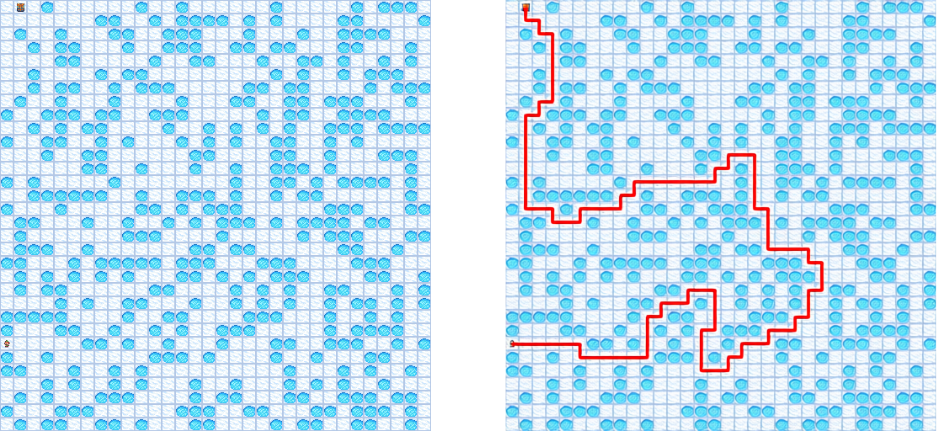} \\ 
    \caption{\textbf{Success case of DiffThinker on VSP-Super.}}
    \label{fig:success2}
\end{figure*}

\begin{figure*}[ht]
    \centering
    \includegraphics[width=0.98\linewidth]{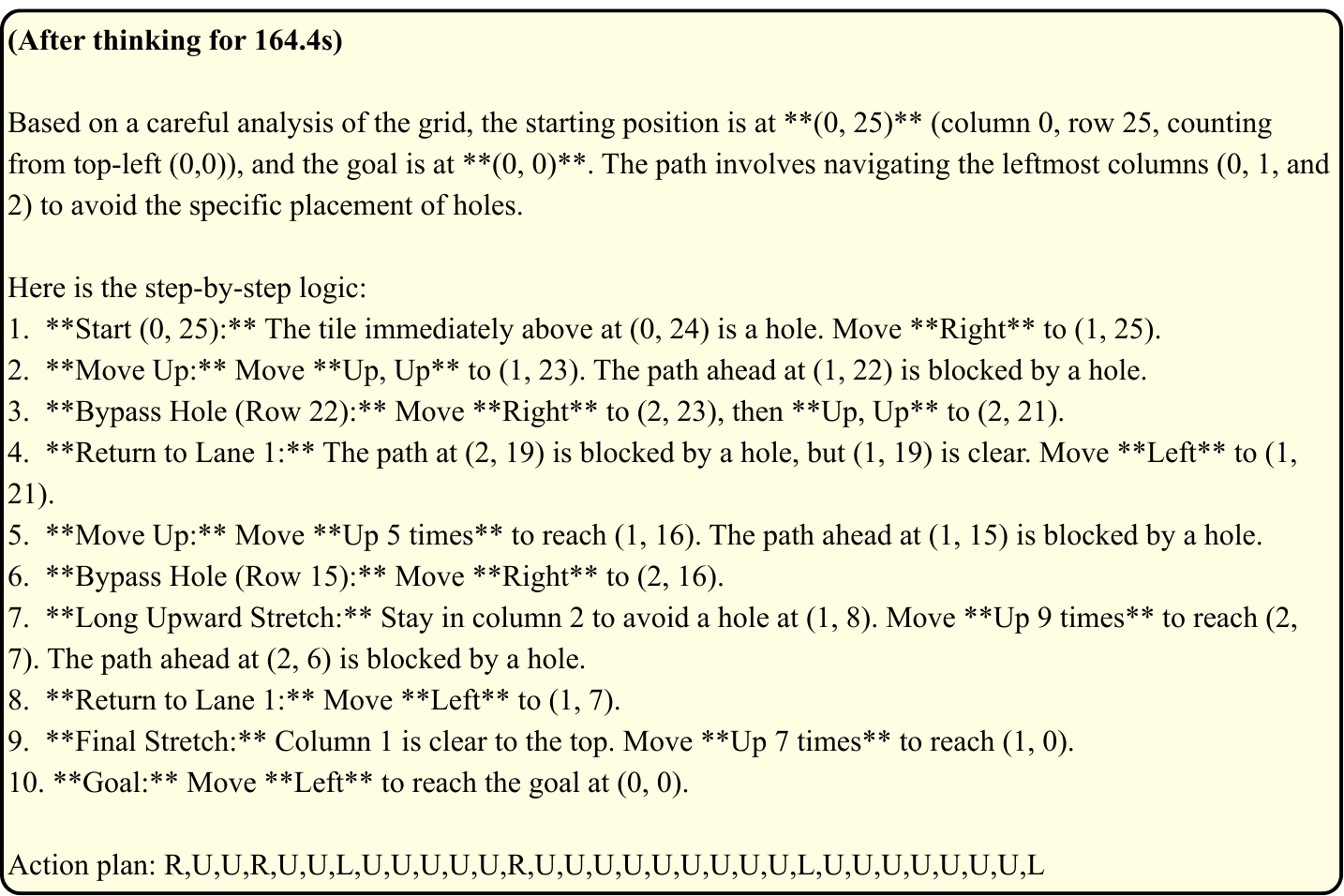} \\ 
    \caption{\textbf{Thinking process of Gemini-3-Pro on VSP-Super.} Gemini-3-Pro fails to provide the correct solution.}
    \label{fig:mllm2}
\end{figure*}

\begin{figure*}[ht]
    \centering
    \includegraphics[width=0.98\linewidth]{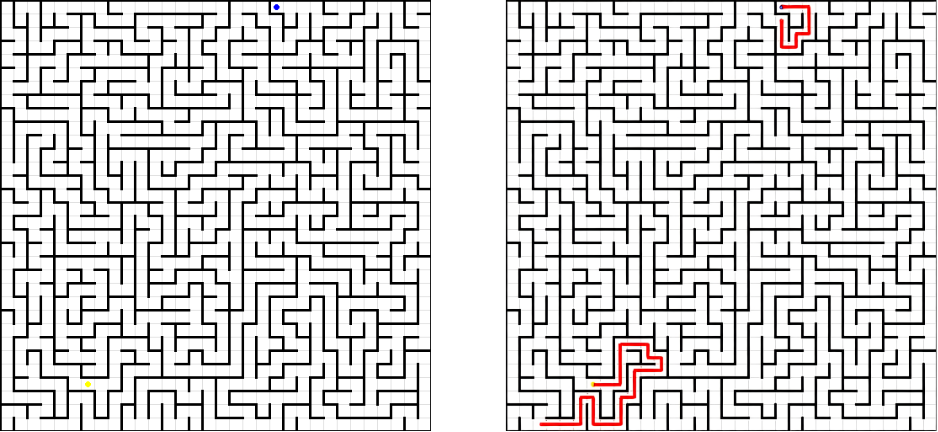} \\ 
    \caption{\textbf{Failure case of DiffThinker on Maze.} In an instance characterized by a significant distance between the starting point and the goal, DiffThinker fails to sustain deep reasoning and provides only a preliminary trajectory.}
    \label{fig:fail3}
\end{figure*}

\begin{figure*}[ht]
    \centering
    \includegraphics[width=0.98\linewidth]{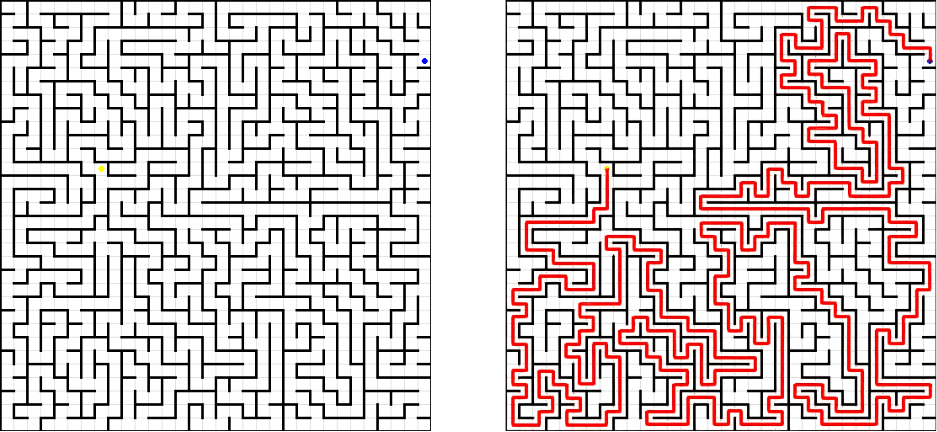} \\ 
    \caption{\textbf{Success case of DiffThinker on Maze.}}
    \label{fig:success3}
\end{figure*}

\begin{figure*}[ht]
    \centering
    \includegraphics[width=0.98\linewidth]{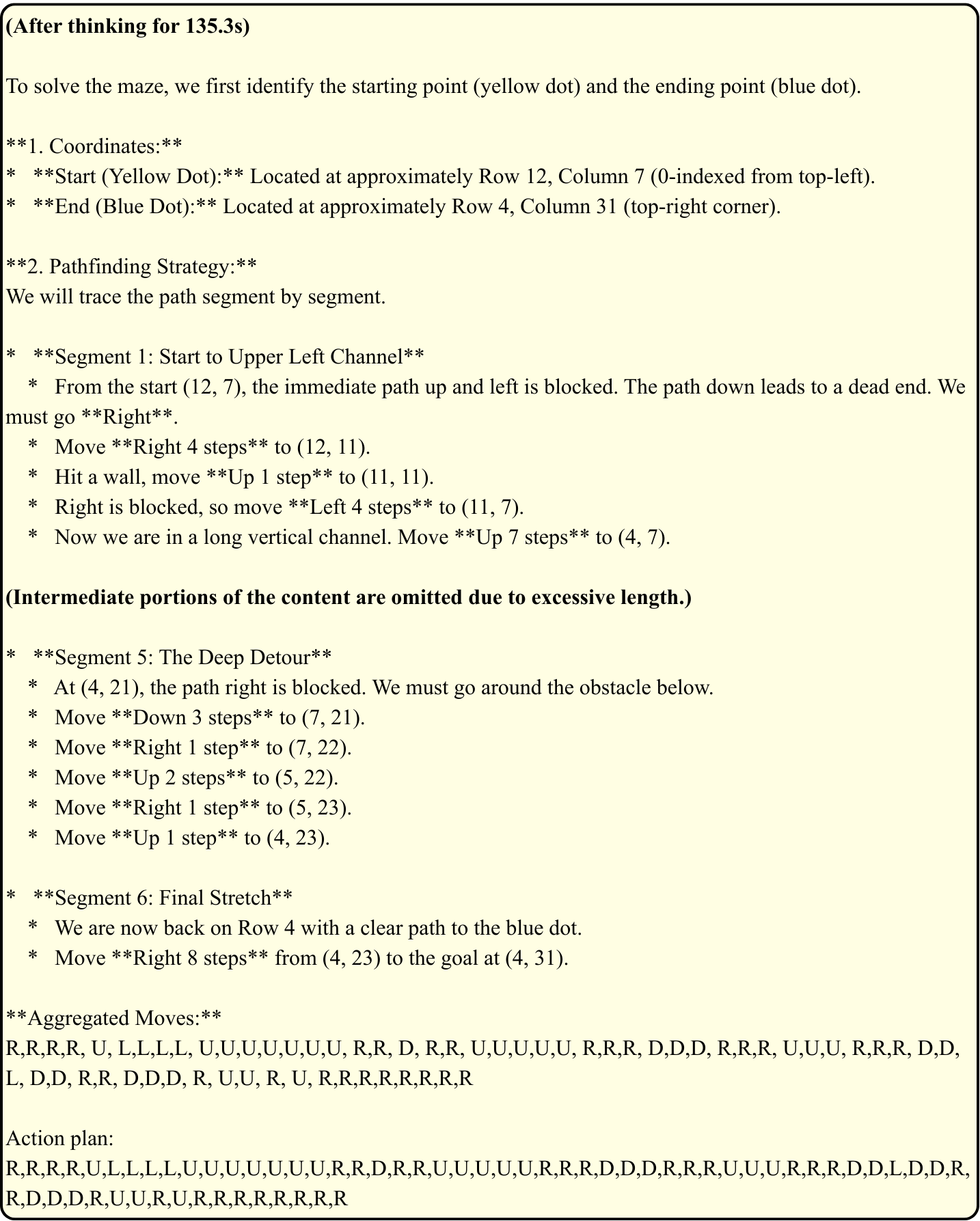} \\ 
    \caption{\textbf{Thinking process of Gemini-3-Pro on Maze.} Gemini-3-Pro fails to provide the correct solution.}
    \label{fig:mllm3}
\end{figure*}

\begin{figure*}[ht]
    \centering
    \includegraphics[width=0.98\linewidth]{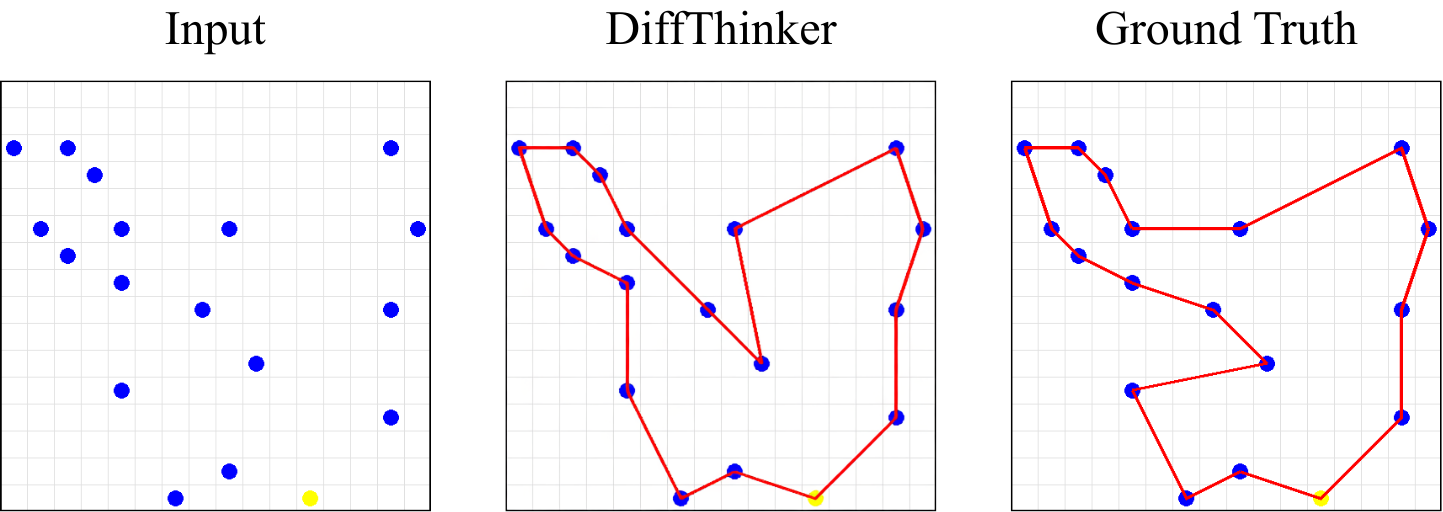} \\ 
    \caption{\textbf{Failure case of DiffThinker on TSP.} DiffThinker successfully identifies a feasible closed loop, yet it is not the shortest path.}
    \label{fig:fail4}
\end{figure*}

\begin{figure*}[ht]
    \centering
    \includegraphics[width=0.98\linewidth]{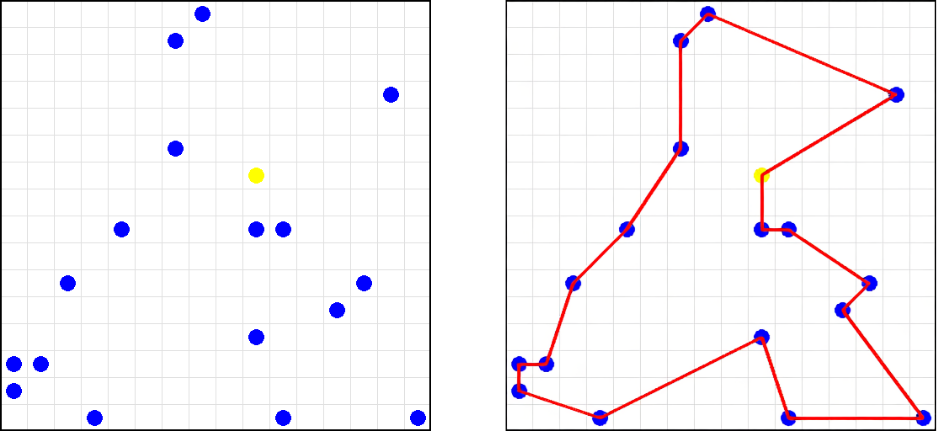} \\ 
    \caption{\textbf{Success case of DiffThinker on TSP.}}
    \label{fig:success4}
\end{figure*}

\begin{figure*}[ht]
    \centering
    \includegraphics[width=0.98\linewidth]{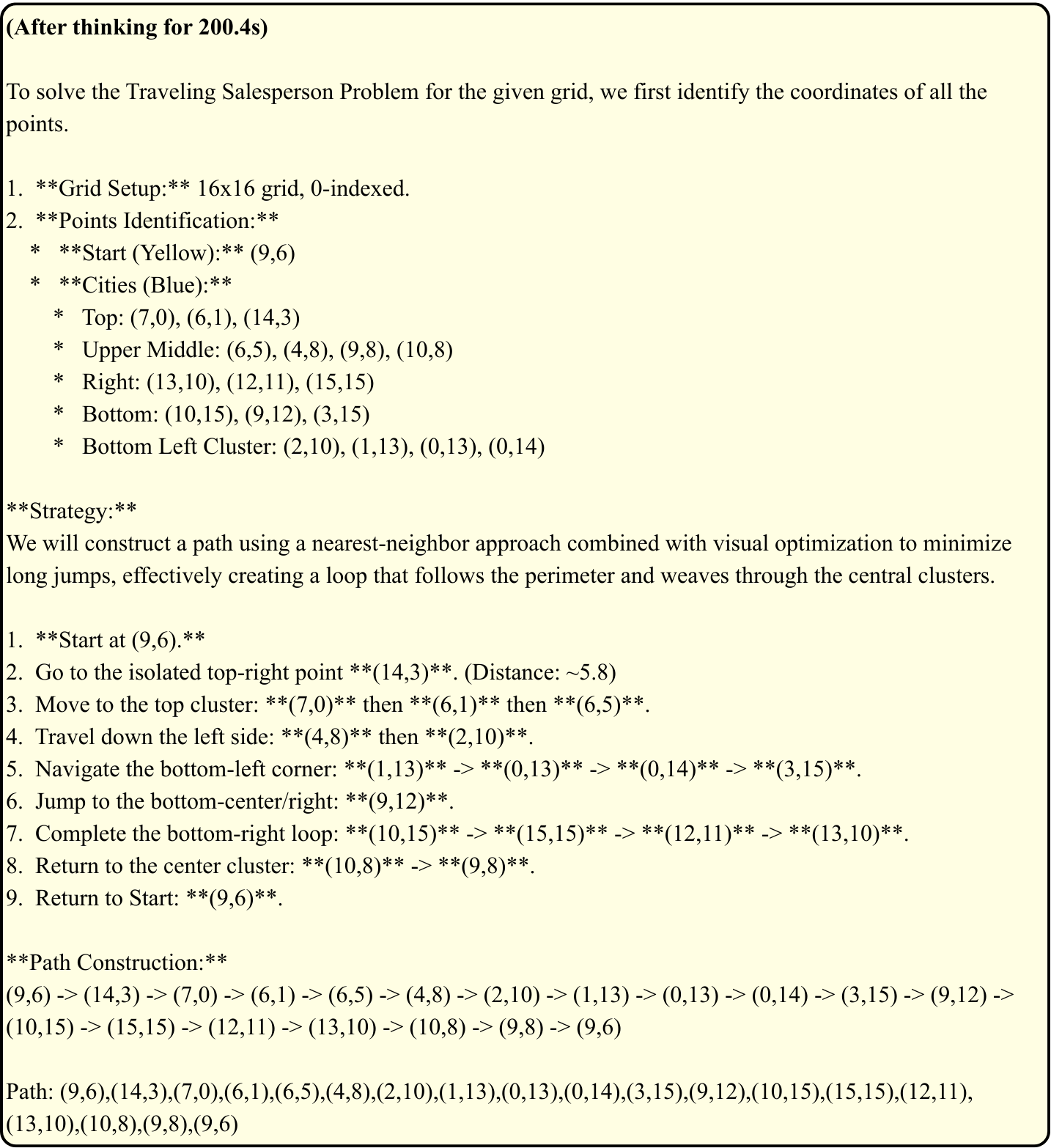} \\ 
    \caption{\textbf{Thinking process of Gemini-3-Pro on TSP.} Gemini-3-Pro successfully provides the correct solution.}
    \label{fig:mllm4}
\end{figure*}

\begin{figure*}[ht]
    \centering
    \includegraphics[width=0.98\linewidth]{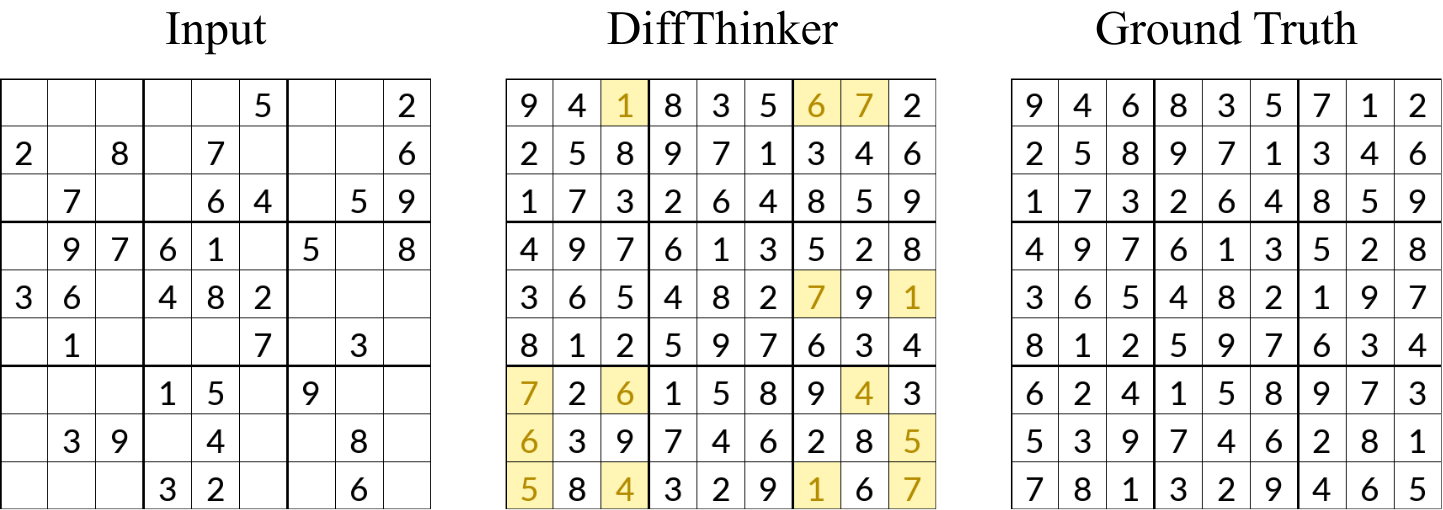} \\ 
    \caption{\textbf{Failure case of DiffThinker on Sudoku.} DiffThinker successfully populates most of entries, yet commits several errors.}
    \label{fig:fail5}
\end{figure*}

\begin{figure*}[ht]
    \centering
    \includegraphics[width=0.98\linewidth]{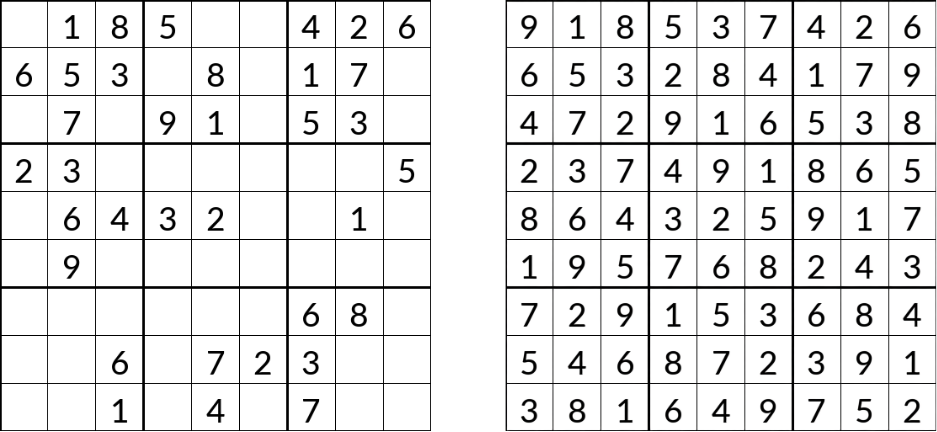} \\ 
    \caption{\textbf{Success case of DiffThinker on Sudoku.}}
    \label{fig:success5}
\end{figure*}

\begin{figure*}[ht]
    \centering
    \includegraphics[width=0.98\linewidth]{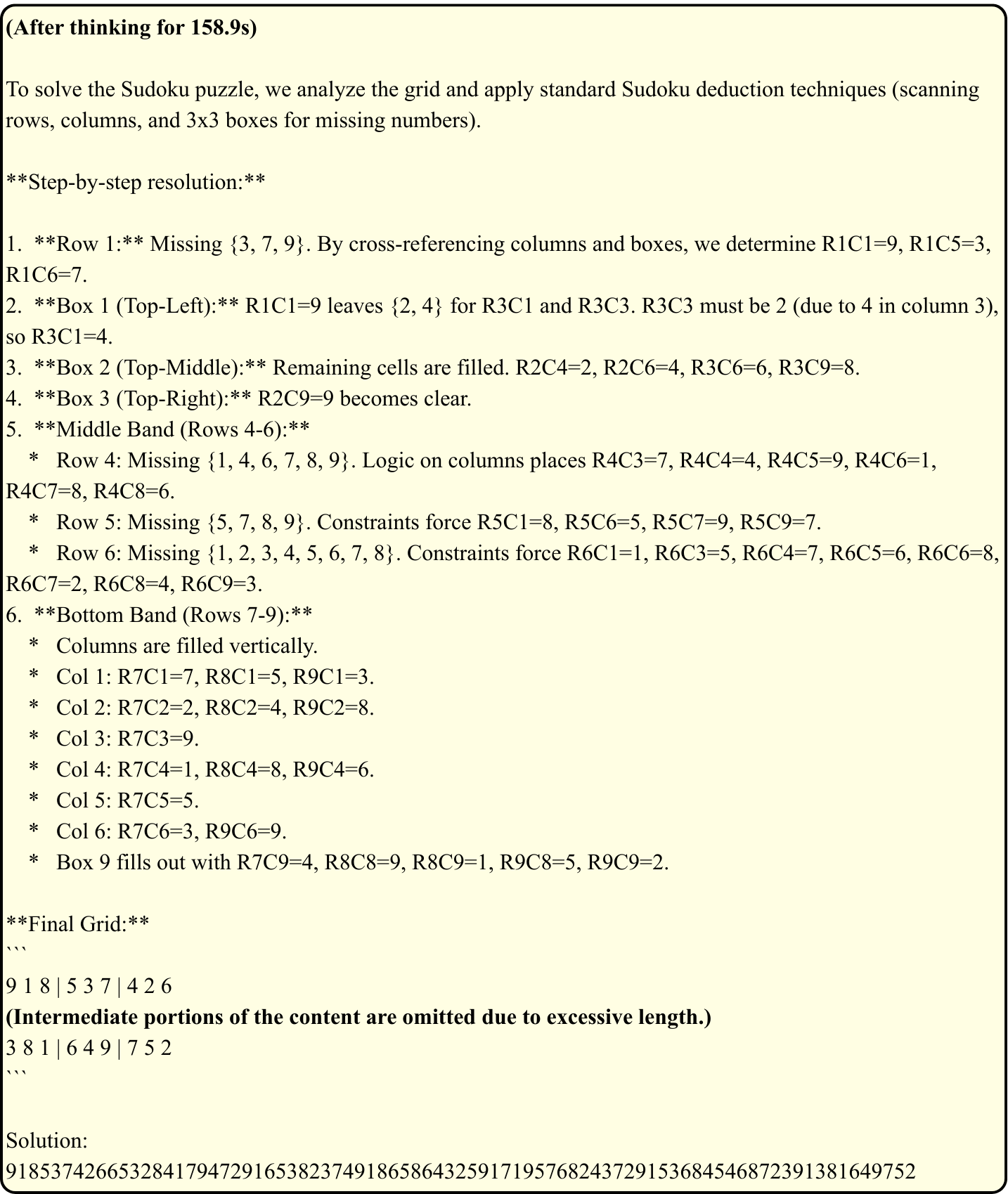} \\ 
    \caption{\textbf{Thinking process of Gemini-3-Pro on Sudoku.} Gemini-3-Pro successfully provides the correct solution.}
    \label{fig:mllm5}
\end{figure*}

\begin{figure*}[ht]
    \centering
    \includegraphics[width=0.98\linewidth]{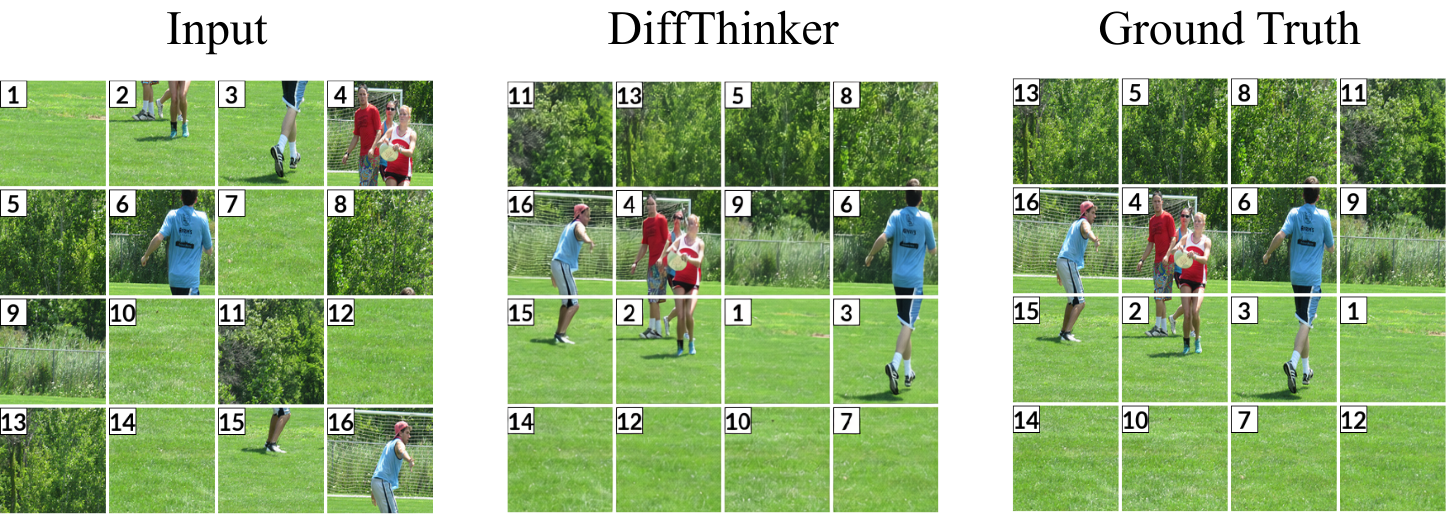} \\ 
    \caption{\textbf{Failure case of DiffThinker on Jigsaw.} Due to the fact that our test set is generated at random, certain instances contain regions that are extremely difficult to distinguish. DiffThinker produces a globally reasonable image, yet the fine details remain incorrect.}
    \label{fig:fail6}
\end{figure*}

\begin{figure*}[ht]
    \centering
    \includegraphics[width=0.98\linewidth]{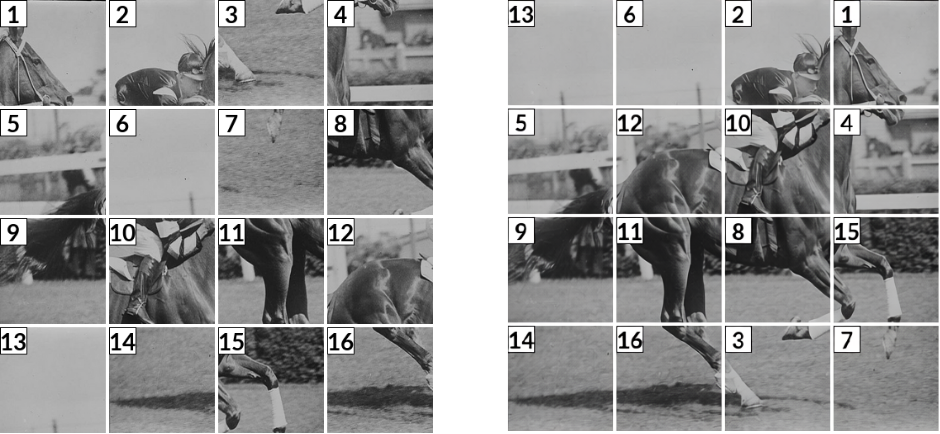} \\ 
    \caption{\textbf{Success case of DiffThinker on Jigsaw.}}
    \label{fig:success6}
\end{figure*}

\begin{figure*}[ht]
    \centering
    \includegraphics[width=0.98\linewidth]{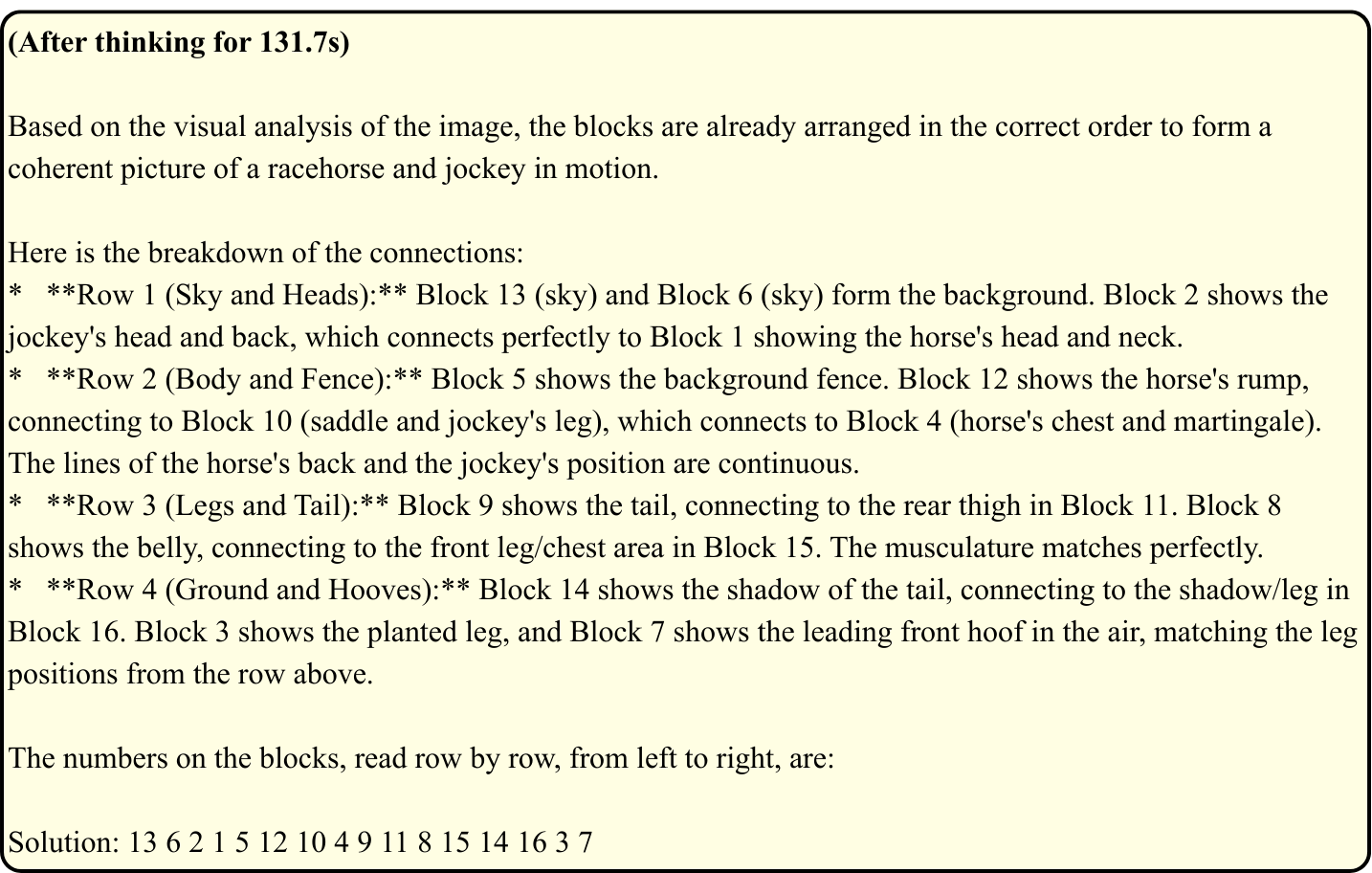} \\ 
    \caption{\textbf{Thinking process of Gemini-3-Pro on Jigsaw.} Gemini-3-Pro successfully provides the correct solution.}
    \label{fig:mllm6}
\end{figure*}

\begin{figure*}[ht]
    \centering
    \includegraphics[width=0.98\linewidth]{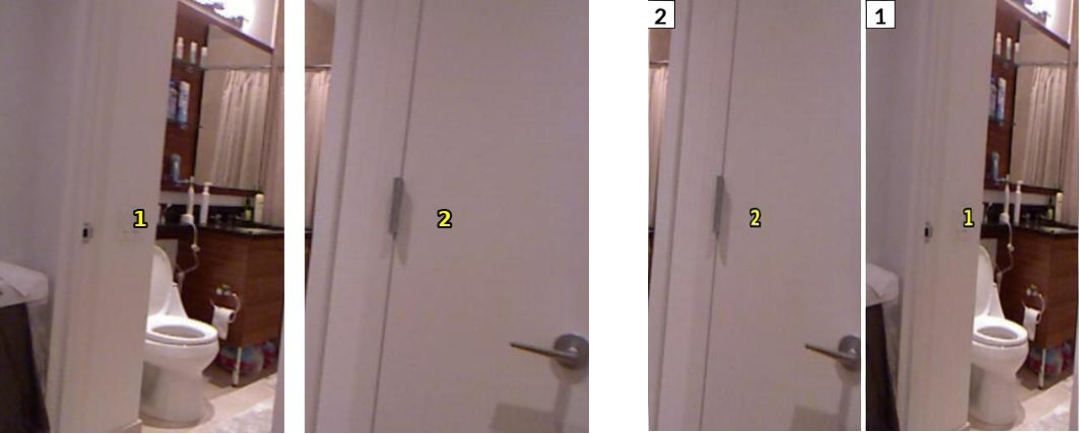} \\ 
    \caption{\textbf{Failure case of DiffThinker on VisPuzzle.}}
    \label{fig:fail7}
\end{figure*}

\begin{figure*}[ht]
    \centering
    \includegraphics[width=0.98\linewidth]{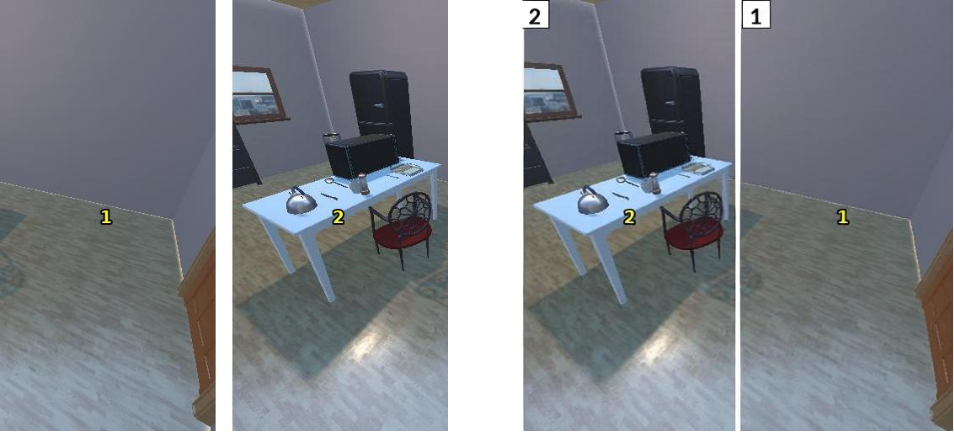} \\ 
    \caption{\textbf{Success case of DiffThinker on VisPuzzle.}}
    \label{fig:success7}
\end{figure*}

\begin{figure*}[ht]
    \centering
    \includegraphics[width=0.98\linewidth]{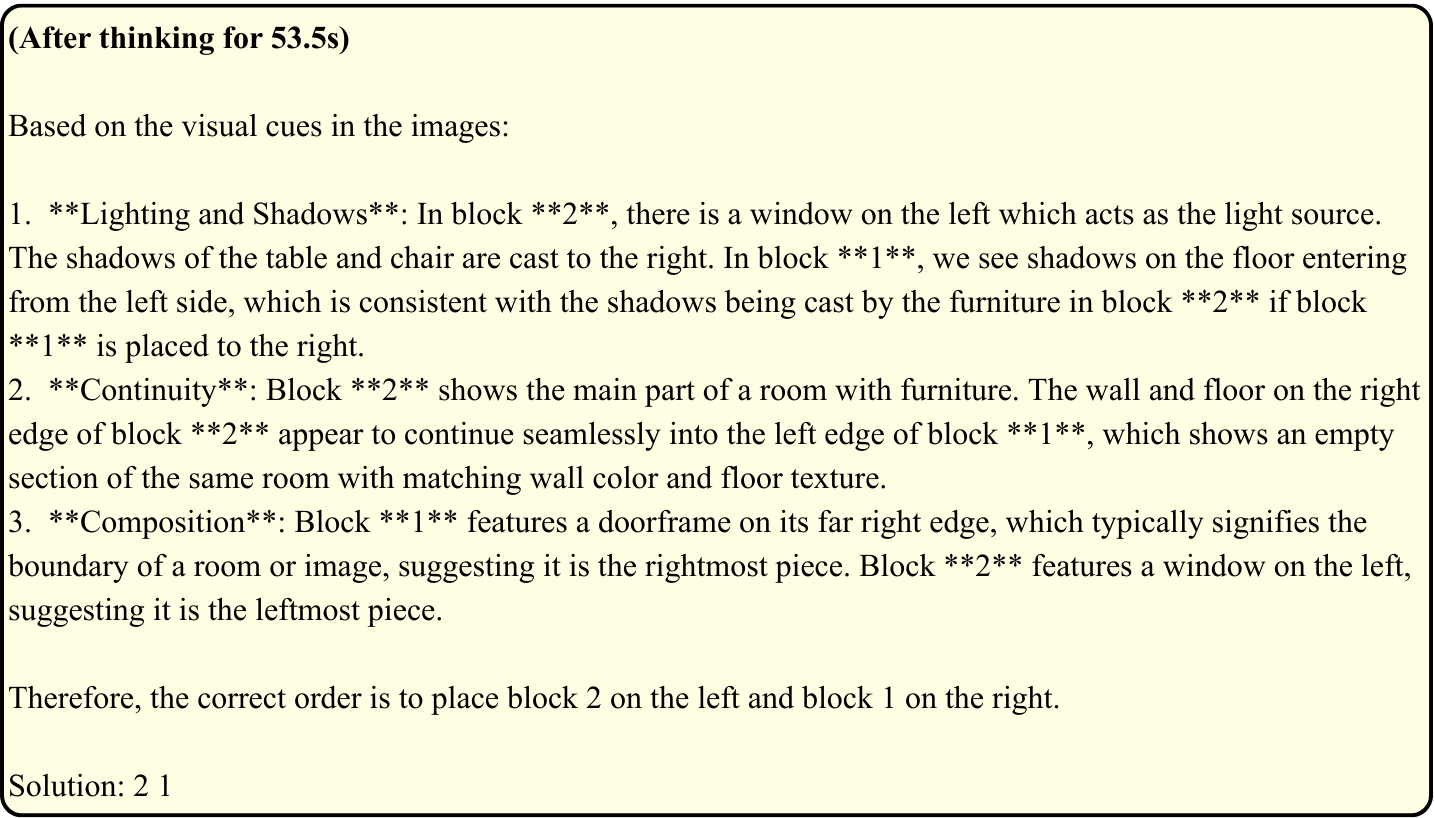} \\ 
    \caption{\textbf{Thinking process of Gemini-3-Pro on VisPuzzle.} Gemini-3-Pro successfully provides the correct solution.}
    \label{fig:mllm7}
\end{figure*}

To facilitate a better understanding of the performance disparities between DiffThinker and MLLMs, we provide success and failure cases of DiffThinker for each task, along with the Thinking processes of Gemini-3-Pro~\cite{gemini3}, as shown in Figures \ref{fig:fail1} through \ref{fig:mllm7}. We utilize Google AI Studio to evaluate Gemini-3-Pro and obtain its reasoning duration. For each task, we evaluate Gemini-3-Pro on the same problem instances where DiffThinker achieved successful solutions.

\end{document}